\def\isarxiv{1} 

\ifdefined\isarxiv
\documentclass[11pt]{article}

\usepackage[numbers]{natbib}

\else

\documentclass[conference]{IEEEtran}
\IEEEoverridecommandlockouts

\def\BibTeX{{\rm B\kern-.05em{\sc i\kern-.025em b}\kern-.08em
    T\kern-.1667em\lower.7ex\hbox{E}\kern-.125emX}}
\fi

\usepackage{amsmath}
\usepackage{amsthm}
\usepackage{amssymb}
\usepackage{algorithm}
\usepackage{subfig}
\usepackage{algpseudocode}
\usepackage{graphicx}
\usepackage{grffile}
\usepackage{wrapfig,epsfig}
\usepackage{url}
\usepackage{xcolor}
\usepackage{epstopdf}

\usepackage{bbm}
\usepackage{dsfont}

\allowdisplaybreaks

\ifdefined\isarxiv

\usepackage{tikz}
\usepackage{hyperref}  
\hypersetup{colorlinks=true,citecolor=blue,linkcolor=blue} 
\usetikzlibrary{arrows}
\usepackage[margin=1in]{geometry}

\else

\usepackage{microtype}
\usepackage{hyperref}
\definecolor{mydarkblue}{rgb}{0,0.08,0.45}
\hypersetup{colorlinks=true, citecolor=mydarkblue,linkcolor=mydarkblue}

\fi
\graphicspath{{./figs/}}

\newtheorem{theorem}{Theorem}[section]
\newtheorem{lemma}[theorem]{Lemma}
\newtheorem{definition}[theorem]{Definition}

\newtheorem{fact}[theorem]{Fact}

\newcommand{\wt}{\widetilde}
\newcommand{\ov}{\overline}

\newcommand{\R}{\mathbb{R}}

\renewcommand{\d}{\mathrm{d}}

\DeclareMathOperator*{\Z}{\mathbb{Z}}

\DeclareMathOperator{\poly}{poly}

\DeclareMathOperator{\nnz}{nnz}

\DeclareMathOperator{\rank}{rank}
\DeclareMathOperator{\diag}{diag}

\DeclareMathOperator{\mat}{mat}

\makeatletter
\newcommand*{\RN}[1]{\expandafter\@slowromancap\romannumeral #1@}
\makeatother

\usepackage{lineno}

\begin{document}

\ifdefined\isarxiv

\date{}

\title{A Unified Scheme of ResNet and Softmax}
\author{
Zhao Song\thanks{\texttt{zsong@adobe.com}. Adobe Research.}
\and
Weixin Wang\thanks{\texttt{wwang176@jh.edu}. Johns Hopkins University.}
\and
Junze Yin\thanks{\texttt{junze@bu.edu}. Boston University.}
}

\else

\title{A Unified Scheme of ResNet and Softmax}

\author{
\IEEEauthorblockN{Zhao Song}
\IEEEauthorblockA{
\textit{Adobe Research}\\
\textit{Adobe}\\
San Jose, CA \\
zsong@adobe.com}
\and
\IEEEauthorblockN{Weixin Wang}
\IEEEauthorblockA{\textit{Department of Engineering} \\
\textit{Johns Hopkins University}\\
Baltimore, MD \\
wwang176@jh.edu}
\and 
\IEEEauthorblockN{Junze Yin}
\IEEEauthorblockA{
\textit{Department of Mathematics and Statistics}\\
\textit{Boston University}\\
Boston, MA \\
junze@bu.edu}
}
\maketitle 
\fi

\ifdefined\isarxiv
\begin{titlepage}
  \maketitle
  \begin{abstract}
Large language models (LLMs) have brought significant changes to human society. Softmax regression and residual neural networks (ResNet) are two important techniques in deep learning: they not only serve as significant theoretical components supporting the functionality of LLMs but also are related to many other machine learning and theoretical computer science fields, including but not limited to image classification, object detection, semantic segmentation, and tensors. 

Previous research works studied these two concepts separately. In this paper, we provide a theoretical analysis of the regression problem:
\begin{align*}
\| \langle \exp(Ax) + A x , {\bf 1}_n \rangle^{-1} ( \exp(Ax) + Ax ) - b \|_2^2,
\end{align*}
where $A$ is a matrix in $\mathbb{R}^{n \times d}$, $b$ is a vector in $\mathbb{R}^n$, and ${\bf 1}_n$ is the $n$-dimensional vector whose entries are all $1$. This regression problem is a unified scheme that combines softmax regression and ResNet, which has never been done before. We derive the gradient, Hessian, and Lipschitz properties of the loss function. The Hessian is shown to be positive semidefinite, and its structure is characterized as the sum of a low-rank matrix and a diagonal matrix. This enables an efficient approximate Newton method. 

As a result, this unified scheme helps to connect two previously thought unrelated fields and provides novel insight into loss landscape and optimization for emerging over-parameterized neural networks, which is meaningful for future research in deep learning models.

  \end{abstract}
  \thispagestyle{empty}
\end{titlepage}

{\hypersetup{linkcolor=black}
\tableofcontents
}
\newpage

\else

\begin{abstract}
Large language models (LLMs) have brought significant changes to human society. Softmax regression and residual neural networks (ResNet) are two important techniques in deep learning: they not only serve as significant theoretical components supporting the functionality of LLMs but also are related to many other machine learning and theoretical computer science fields, including but not limited to image classification, object detection, semantic segmentation, and tensors. 

Previous research works studied these two concepts separately. In this paper, we provide a theoretical analysis of the regression problem:
\begin{align*}
\| \langle \exp(Ax) + A x , {\bf 1}_n \rangle^{-1} ( \exp(Ax) + Ax ) - b \|_2^2,
\end{align*}
where $A$ is a matrix in $\mathbb{R}^{n \times d}$, $b$ is a vector in $\mathbb{R}^n$, and ${\bf 1}_n$ is the $n$-dimensional vector whose entries are all $1$. This regression problem is a unified scheme that combines softmax regression and ResNet, which has never been done before. We derive the gradient, Hessian, and Lipschitz properties of the loss function. The Hessian is shown to be positive semidefinite, and its structure is characterized as the sum of a low-rank matrix and a diagonal matrix. This enables an efficient approximate Newton method. 

As a result, this unified scheme helps to connect two previously thought unrelated fields and provides novel insight into loss landscape and optimization for emerging over-parameterized neural networks, which is meaningful for future research in deep learning models.

\end{abstract}

\fi

\section{Introduction}

Softmax regression and residual neural networks (ResNet) are two emerging techniques in deep learning that have driven advances in computer vision and natural language processing tasks. In previous research, these two methods were studied separately.  

\begin{definition}[Softmax regression, \cite{dls23}]\label{def:softmax_regression}
     Given a matrix $A \in \R^{n \times d}$ and a vector $b \in \R^n$, the goal of the softmax regression is to compute the following problem:
    \begin{align*}
        \min_{x \in \R^d} \| \langle \exp(Ax) , {\bf 1}_n \rangle^{-1} \exp(Ax) - b \|_2^2,
    \end{align*}
    where ${\bf 1}_n$ denotes the $n$-dimensional vector whose entries are all $1$.
\end{definition}

\begin{figure}[!ht]
    \centering
    \includegraphics[width = \linewidth]{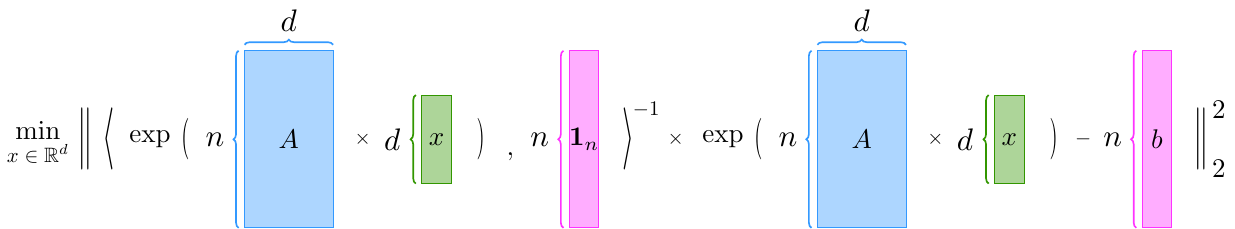}
    \caption{The visualization of the softmax regression (see Definition~\ref{def:softmax_regression}). $A$ is a matrix in $\R^{n \times d}$. $b, {\bf 1}_n \in \R^n$ and $x \in \R^d$ are vectors. First, we compute $\exp(Ax)$ by multiplying $A$ with $x$ and then calculating the exponential of each entry in their product. Second, we compute the inner product of $\exp(Ax)$ and ${\bf 1}_n$ and then find the multiplicative inverse of it. Third, we multiply the results of the first and the second step and subtract $b$ from it. Finally, we compute the minimum of the square of the $\ell_2$ norm of result of the third step. The blue rectangles represent the $n \times d$ matrices. The pink rectangles represent the $n$-dimensional vectors. The green rectangles represent the $d$-dimensional vectors.}
    \label{fig:softmax_regression}
\end{figure}

Because of the explosive development of large language models (LLMs), there is an increasing amount of work focusing on the theoretical aspect of LLMs, aiming to improve the ability of LLMs from different aspects, including sentiment analysis \cite{uas+20}, natural language translation \cite{hwl21}, creative writing \cite{o22, o23}, and language modeling \cite{mms+19}. One of the most important components of an LLM is its ability to identify and focus on the relevant information from the input text. Theoretical works \cite{gsy23_dp,lsz23,dls23,bsz23,gsy23_hyper,gms23,as23,zhdk23} analyze the attention computation to support this ability. 
\begin{definition}[Attention computation]
    Let $Q$, $K$, and $V$ be $n \times d$ matrices whose entries are all real numbers. 

    Let $A = \exp(QK^\top)$ and $D = \diag(A {\bf 1}_n)$ be $n$-dimensional square matrices, where $\diag(A {\bf 1}_n)$ is a diagonal matrix whose entries on the $i$-th row and $i$-th column is the same as the $i$-th entry of the vector $A {\bf 1}_n$.
 
    The static attention computation is defined as 
    \begin{align*}
        \mathsf{Att}(Q, K, V ) := D^{-1} AV.
    \end{align*}
\end{definition}

In attention computation, the matrix $Q$ is denoted as the query tokens, which are derived from the previous hidden state of decoders. $K$ and $V$ represent the key tokens and values. When computing $A$, the softmax function is applied to get the attention weight, namely $A_{i, j}$. Inspired by the role of the exponential functions in attention computation, prior research \cite{gms23,lsz23} has built a theoretical framework of hyperbolic function regression, which includes the functions $f(x) = \exp(Ax), \cosh(Ax)$, and $\sinh(Ax)$.

\begin{definition}[Hyperbolic regression, \cite{lsz23}]\label{def:hyperbolic_regression}
    Given a matrix $A \in \R^{n \times d}$ and a vector $b \in \R^n$, the goal of the hyperbolic regression problem is to compute the following regression problem:
    \begin{align*}
        \min_{x \in \R^d} \| f(x) - b \|_2.
    \end{align*}
\end{definition}

The approach developed by \cite{dls23} for analyzing the hyperbolic regression is to consider the normalization factor, namely $\langle f(x), {\bf 1}_n \rangle^{-1} = \langle \exp(Ax), {\bf 1}_n \rangle^{-1}$. By focusing on the $\exp$, \cite{dls23} transform the hyperbolic regression problem (see Definition~\ref{def:hyperbolic_regression}) to the softmax regression problem (see Definition~\ref{def:softmax_regression}). Later on, \cite{lsx+23} studies the in-context learning based on a softmax regression of attention mechanism in the Transformer, which is an essential component within LLMs since it allows the model to focus on particular input elements. Moreover, \cite{gsx23} utilize a tensor-trick from \cite{szz21,z22,djs+19,swyz21,swz19,dssw18} simplifying the multiple softmax regression into a single softmax regression.

ResNet is a certain type of deep learning model: the weight layers can learn the residual functions \cite{hzrs16}. It is characterized by skip connections, which may perform identity mappings by adding the layer's output to the initial input. This mechanism is similar to the Highway Network in \cite{sgs15} that the gates are opened through highly positive bias weights. This innovation facilitates the training of deep learning models with a substantial number of layers, allowing them to achieve better accuracy as they become deeper. These identity skip connections, commonly known as ``residual connections", are also employed in various other systems, including Transformer \cite{vsp+17}, BERT \cite{dclt18}, and ChatGPT \cite{o22}. Moreover, ResNets have achieved state-of-the-art performance across many computer vision tasks, including image classification \cite{mc19,spba21}, object detection \cite{oyz+19,lknr19,llgz19,hlk19}, and semantic segmentation \cite{fef+17,xyz19,wcy+18,dzl+21}. Mathematically, it is defined as
\begin{align}\label{eq:resnet}
    Y_{j+1} = Y_j + F(Y_j , \theta_j )
\end{align}
where $Y_{j+1}, Y_j, F(Y_j , \theta_j ) \in \R^d$: $Y_j$ represents the feature values at the $j$-th layer, while $\theta_j$ denotes the network parameters specific to that layer. The objective of the training process is to learn the network parameters $\theta$.

In this paper, we combine the softmax regression (see Definition~\ref{def:softmax_regression}) with ResNet and give a theoretical analysis of this problem. We formally define it as follows:
\begin{definition}[Soft-Residual Regression]\label{def:our_regression}
Given a matrix $A \in \R^{n \times d}$ and a vector $b \in \R^n$, the goal is to compute the following regression problem:
\begin{align*}
\| \langle \exp(Ax) + A x , {\bf 1}_n \rangle^{-1} ( \exp(Ax) + Ax ) - b \|_2^2
\end{align*}

\end{definition}

\begin{figure}[!ht]
    \centering
    \includegraphics[width = \linewidth]{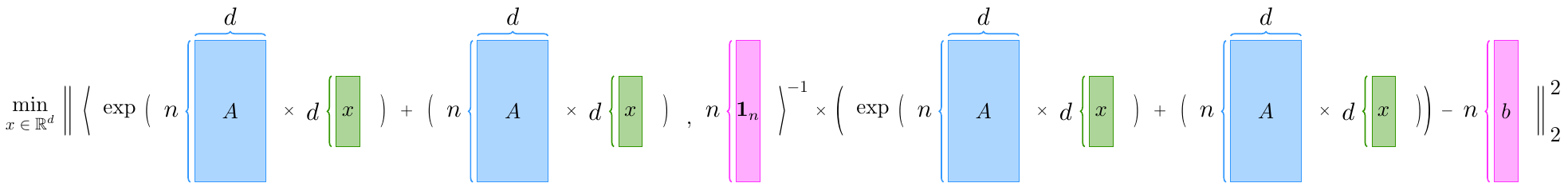}
    \caption{The visualization of the soft-residual regression (see Definition~\ref{def:our_regression}). $A$ is a matrix in $\R^{n \times d}$. $b, {\bf 1}_n \in \R^n$ and $x \in \R^d$ are vectors. First, we compute $\exp(Ax)$ by multiplying $A$ with $x$ and then calculating the exponential of each entry in their product. Second, we add $\exp(Ax)$ with $Ax$. Third, we compute the inner product of $\exp(Ax) + Ax$ and ${\bf 1}_n$ and then find the multiplicative inverse of it. Fourth, we multiply the results of the third step with $\exp(Ax) + Ax$ and subtract $b$ from it. Finally, we compute the minimum of the square of the $\ell_2$ norm of the result of the fourth step. The blue rectangles represent the $n \times d$ matrices. The pink rectangles represent the $n$-dimensional vectors. The green rectangles represent the $d$-dimensional vectors.}
    \label{fig:soft_residual_regression}
\end{figure}

We are motivated by the fact that the softmax regression and ResNets have mostly been studied separately in prior works. We would like to provide a theoretical analysis for combining them together. The unified perspective and analysis of the loss landscape could provide insights into optimization and generalization for emerging overparametrized models. We firmly believe that this lays the groundwork for further research at the intersection of softmax classification and residual architectures.

\paragraph{Roadmap}

In Section~\ref{sec:related_work}, we introduce the related work. In Section~\ref{sec:preli}, we introduce the basic notations we use and present basic mathematical facts that may support the mathematical properties developed in this paper. In Section~\ref{sec:gradient}, we compute the gradient of the functions we defined earlier. In Section~\ref{sec:hessian}, we compute the Hessian of these functions based on their gradient. In Section~\ref{sec:rewrite_hessian}, we formally define the key functions that appear in the Hessian and rewrite the Hessian functions in a more formal way. In Section~\ref{sec:hessian_PSD}, we show that the Hessian is positive semidefinite (PSD). In Section~\ref{sec:hessian_lip}, we show that the Hessian is Lipschitz. In Section~\ref{sec:main_result}, we summarize the mathematical properties we developed in the previous sections and explain how they may support the main result of this paper. In Section~\ref{sec:conclusion}, we conclude this paper, present the meaningfulness of our work, and discuss future research directions. 

\section{Related Work}
\label{sec:related_work}

In this section, we introduce the previous related research works.

\paragraph{Residual Neural Networks}

ResNets were introduced by \cite{hzrs16} for the purpose of simplifying the training of networks that are significantly deeper than those that were used previously. Its design is inspired by residual learning. \cite{hzrs16} demonstrated state-of-the-art performance on image recognition benchmarks using extremely deep ResNets with over $100$ layers. By adding shortcut connections, ResNets were able to successfully train far deeper networks than previous architectures. 

After being introduced, ResNets have become a prevalent research area in computer vision and its application. Many subsequent works were built based on the original ResNet model. In \cite{xgd+17}, ResNeXt is proposed: it splits each layer into smaller groups to increase the cardinality, which is defined as the size of the set of transformations. It is shown that the increase in cardinality leads to higher classification accuracy and is more effective than going deeper and wider when the capacity is increased. Moreover, \cite{zk16} propose a novel architecture called wide residual networks (WRNs), and based on their experimental study, it is shown that residual networks with a reduced number of layers and an increased number of the network's width are far superior to the thin and deep counterparts. In addition, ResNets is also studied with efficiency \cite{hzc+17,lhww20,rel19}, video analysis \cite{twt+18,psb+22,zql+22,aac+20,ak21}, and breast cancer \cite{vrd+18,ijf+22,ag22,l23,mma23}. 

Moreover, ResNets has been found a connection with ordinary differential equations (ODEs). ResNets, as shown in Eq.~\eqref{eq:resnet}, is a difference equation (or a discrete dynamical system). ODEs consider the continuous change of a dynamical system with respect to time. Therefore, a small parameter $h > 0$ is introduced in Eq.~\eqref{eq:resnet}, which makes this equation become continuous:
\begin{align*}
    \frac{Y_{j+1} - Y_j}{h} = F(Y_j , \theta_j ),
\end{align*}
which implies
\begin{align*}
    \frac{\d Y(t)}{\d t} \approx F(Y(t), \theta(t)), ~ Y(0) = Y_0.
\end{align*}

Due to the lack of general guidance to network architecture design, \cite{lzld18} connect the concept of ResNets with numerical differential equations, showing that ResNet can be interpreted as forward Euler discretizations of differential equations. Follow-up works expand this connection: \cite{crbd18} improves the accuracy by introducing more stable and adaptive ODE solvers for the use of ResNets, \cite{hr17} establishes a new architecture based on ODEs to resolve the challenge of the vanishing gradient, and \cite{cmh+18} construct a theoretical framework studying stability and reversibility of deep neural networks: there are three reversible neural network architectures that are developed, which theoretically can go arbitrary deep.

\paragraph{Attention}

The attention matrix is a square matrix, which contains the associations between words or tokens in natural language text. Each row and column of an attention matrix align with the corresponding token, and the values within it signify the level of connection between these tokens. When generating the output, the attention matrix has a huge influence on determining the significance of individual input tokens within a sequence. Under this attention mechanism, each input token is assigned a weight or score that reflects its relevance with the current output generation. 

There are various methods that have been developed to approximate the prominent entries of the attention matrix: methods like k-means clustering \cite{dkod20} and Locality Sensitive Hashing (LSH) \cite{syy21,clp+21,kkl20} are to restrict the attention to nearby tokens, and other methods like \cite{cld+20} approximate the attention matrix by using the random feature maps based on Gaussian or exponential kernels. Furthermore, \cite{cdw+21} presented that combining LSH-based and random feature-based methods is a more effective technique for estimating the attention matrix.

As presented in the recent works \cite{gsy23_dp,bsz23,zhdk23,as23,gms23,gsy23_hyper,lsz23,dls23} the calculation of inner product attention is a critical task. It is essential in training LLMs, like Transformer \cite{vsp+17}, GPT-1 \cite{rns+18}, BERT \cite{dclt18}, GPT-2 \cite{rwc+19}, GPT-3 \cite{bmr+20}, and ChatGPT, all of which have shown the extraordinary performance in handling natural language processing tasks, compared to smaller language models and traditional algorithms. Various studies have delved into different aspects of attention computation, such as softmax regression exploration \cite{gms23,dls23}, exponential regression analysis \cite{lsz23}, algorithms and complexity analysis for static attention computation \cite{as23}, private computation of the attention matrix \cite{gsy23_dp}, and maintaining the attention matrix dynamically \cite{bsz23}. Additionally, there's an algorithm for rescaled softmax regression \cite{gsy23_hyper}, which presents an alternative formulation compared to exponential \cite{lsz23} and softmax regression methods \cite{dls23}. \cite{syz23} studies the attention kernel regression problem through the pre-conditioner. \cite{gswy23} studies the single layer of attention via the tensor and SVM tricks.

\paragraph{Softmax Regression}

The softmax unit is a fundamental component of LLMs and serves a vital purpose: it enables the model to create a probability distribution concerning the possible following words or phrases when presented with a sequence of input words. Additionally, the softmax unit may allow the models to adapt their neural network's weights and biases based on the available data. Under convex optimization, the softmax function is utilized for managing the progress and stability of potential functions, as shown in \cite{b20,cls21}. Drawing inspiration from the concept of the softmax unit, \cite{dls23} introduces a problem known as softmax regression. The studies study three particular formulations: exponential regression \cite{gms23,lsz23}, softmax regression \cite{dls23,lsx+23,ssz23,wyw+23,zsz+23}, the rescaled softmax regression \cite{gsy23_hyper}, and multiple softmax regression \cite{gsx23}.

\paragraph{Convergence and Optimization}

There are numerous studies analyzing the optimization and convergence to enhance training methods. \cite{ll18} reveals that stochastic gradient descent may efficiently optimize over-parameterized neural networks for the structured data. Similarly, \cite{dzps18} shows that the gradient descent can also optimize over-parameterized neural networks. After that, \cite{azls19a} introduces a convergence theory for over-parameterized deep neural networks using gradient descent. Meanwhile, \cite{azls19b} investigates the rate at which training recurrent neural networks converge.

\cite{adh+19a} gives an in-depth analysis of the optimization and generalization of over-parameterized two-layer neural networks. Moreover, \cite{adh+19b} analyzes the exact computation using infinitely wide neural networks. On the other hand, \cite{cgh+19} proposes the Gram-Gauss-Newton method, which is used to optimize over-parameterized neural networks.

In \cite{zg19}, global convergence of stochastic gradient descent is analyzed during the training of deep neural networks, which requires less over-parameterization compared to previous research. Furthermore, the works like \cite{zpd+20,jt19,os20} focus on the optimization and generalization aspects, whereas \cite{lsz23,gms23} emphasize the convergence rate and stability.

Moreover, there are works such as \cite{z22,als+22,bpsw20,mosw22,szz21} that concentrate on specialized optimization algorithms and techniques for training neural networks. Finally, \cite{hlsy21,lss+20} centers their efforts on harnessing the structural aspects of neural networks for specific purposes.

In addition, there is a significant amount of work \cite{syyz23_dp,swyz23,swyz21,qszz23,rsz22,lsz+23,qrs+22,qjs+22,qsz23,qsw23,qsy23,syyz23,dsy23,syyz23_linf,gsy23_coin,gsyz23} that analyze sketching: a technique to speed up machine learning algorithms
and optimization.

\section{Preliminary}
\label{sec:preli}

In this section, we first introduce the basic notations we use. Then, in section~\ref{sub:preli:badic_def}, we introduce the definition of the functions we analyze in the later sections. In Section~\ref{sub:preli:badic_facts}, we present the basic mathematical properties of the derivative, vectors, norms, and matrices.  

\paragraph{Notations}

Now, we define the notations used in this paper. 

First, we define the notations related to sets. Let $\Z_+$ be the set containing all the positive integers, namely $\{1, 2, 3, \dots\}$. Let $n, d$ be arbitrary elements in $\Z_+$. We define $[n] := \{1, 2, \dots, n\}$. We define $\R, \R^n, \R^{n \times d}$ to be the set containing all real numbers, the set containing all $n$-dimensional vectors whose entries are all real numbers, and the set containing all $n \times d$ matrices whose entries are all real numbers, respectively. 

Then, we define the notations related to vectors. Let $x, y$ be arbitrary elements in $\R^n$. We use $x_i$ to denote the $i$-th entry of $x$, for all $i \in [n]$. $\| x \|_2 \in \R$ denotes the $\ell_2$ norm of the vector $x$, which is defined as $\| x \|_2 := (\sum_{i=1}^n x_i^2)^{1/2}$. $\langle x, y\rangle \in \R$ represents the inner product of $x$ and $y$, which is defined as $\langle x, y\rangle := \sum_{i=1}^n x_i y_i$. We use $\circ$ to denote a binary operation between $x$ and $y$, called the Hadamard product. $x \circ y \in \R^n$ is defined as $(x \circ y)_i := x_i \cdot y_i$, for all $i \in [n]$. ${\bf 1}_n \in \R^n$ denotes a vector, where $({\bf 1}_n)_i := 1$ for all $i \in [n]$, and ${\bf 0}_n \in \R^n$ denotes a vector, where $({\bf 0}_n)_i := 0$ for all $i \in [n]$.

After that, we introduce the notations related to matrices. Let $A$ be an arbitrary element in $\R^{n \times d}$. We use $A_{i, j}$ to denote the entry of $A$ which is at the $i$-th row and $j$-th column, for all $i \in [n]$ and $j \in [d]$. We define $A_{*, i} \in \R^n$ as $(A_{*, i})_j : = A_{j, i}$, for all $j \in [n]$ and $i \in [d]$. We use $\| A \|$ to denote the spectral norm of $A$, i.e., $\| A \|:= \max_{x \in \R^d} \| A x \|_2 / \| x \|_2$. This also implies that for any $x \in \R^d$, $\| A x \|_2 \leq \| A \| \cdot \| x \|_2$. For any $x \in \R^d$, we define $\diag(x) \in \R^{d \times d}$ as $(\diag(x))_{i, j} : = x_i$ for all $i = j$ and $(\diag(x))_{i, j} : = 0$ for all $i \neq j$, where $i, j \in [d]$. We use $A^\top \in \R^{d \times n}$ to denote the transpose of $A$, namely $(A^\top)_{i, j} : = A_{j, i}$, for all $i \in [d]$ and $j \in [n]$. We use $I_n$ to denote the $n$-dimensional identity matrix. Let $B$ and $C$ be arbitrary symmetric matrices. We say $B \preceq C$ if, for all vector $x$, we have $x^\top B x \leq x^\top C x$. We say $B$ is positive semidefinite (or $B$ is a PSD matrix), denoted as $B \succeq 0$, if, for all vectors $x$, we have $x^\top B x \geq 0$.

Finally, we define the notations related to functions. We define $\phi: \R \to \R$ as $\phi(z) := \max\{z , 0\}$. For a differentiable function $f$, we use $\frac{\d f}{\d x}$ to denote the derivative of $f$.

\subsection{Basic Definitions}
\label{sub:preli:badic_def}

In this section, we define the basic functions which are analyzed in the later sections. 

\begin{definition}[Basic functions]\label{def:basic_functions}
    Let $A \in \R^{n \times d}$ be an arbitrary matrix. Let $x \in \R^d$ be an arbitrary vector. Let $b \in \R^n$ be a given vector. Let $i \in [d]$ be an arbitrary positive integer. We define the functions 
         $u_1, u_2, u, f, c, z, v_i : \R^d \to \R^n$
    and
        $\alpha, L, \beta_i : \R^d \to \R$
    as
    \begin{align*}
        u_1(x) &:= Ax 
        &u_2(x) &:= \exp(Ax) \\
        u(x) &:= u_1(x) + u_2(x) 
        &\alpha(x) &:= \langle u(x), {\bf 1}_n \rangle\\
        f(x) &:= \alpha(x)^{-1} u(x)
        &c(x) &:= f(x) - b\\
        L(x) &:= 0.5 \|c(x)\|_2^2 
        &z(x) &:= u_2(x) + {\bf 1}_n \\
        v_i(x) &:= (u_2(x) + {\bf 1}_n) \circ A_{*, i} 
        &\beta_i(x) &:= \langle v_i(x), {\bf 1}_n \rangle.
    \end{align*}

\end{definition}

\subsection{Basic Facts}
\label{sub:preli:badic_facts}

In this section, we present the basic mathematical properties which are used to support our analysis in later sections.

\begin{fact}\label{fac:exp_differention1}
Let $f$ be a differentiable function. Then, we have
\begin{itemize}
    \item Part 1. $
    \frac{\d}{\d x}\exp(x) = \exp (x)
    $
    \item  Part 2. For any $ j \neq i $, $\frac{\d}{\d x_i}f(x_j) = 0$
\end{itemize}
\end{fact}

\begin{fact}\label{fac:circ_rules}
    For all vectors $u,v,w \in \R^{n}$, we have 
    \begin{itemize}
        \item $\langle u,v \rangle = \langle u \circ v, {\bf 1}_n \rangle =  u^\top \mathrm{diag}(v)  {\bf 1}_n $
        \item $\langle u \circ v, w \rangle = \langle u \circ w, v \rangle$
        \item $\langle u \circ v, w \rangle =  \langle u \circ v \circ w, {\bf 1}_n  \rangle = u^\top \diag(v) w$
        \item $\langle u \circ v \circ w \circ z , {\bf 1}_n \rangle = u^\top \diag(v \circ w) z$
        \item $u \circ  v = v \circ u = \diag (u) \cdot v = \diag (v) \cdot u$ 
        \item $u^{\top}(v \circ w) = v^{\top}(u \circ w) = w^{\top}(u \circ v)= u^{\top}\diag(v) w = v^{\top}\diag(u) w = w^{\top}\diag(u) v$
        \item $\diag (u) \cdot \diag (v) \cdot {\bf 1}_n = \diag(u) v$
        \item $\diag (u \circ v) = \diag (u) \diag (v)$
        \item $\diag (u) + \diag (v) = \diag (u +v)$
        \item $\langle u,v \rangle = \langle v,u \rangle$
        \item $\langle u,v \rangle = u^\top v = v^\top u$ 
        \item $u + v w^\top a = u+v u^\top w = (I_n + vw^\top) u $
        \item $u + v^\top w u = (1+ v^\top w)u $
        
    \end{itemize}
\end{fact}

\begin{fact} \label{fac:exponential_der_rule}
Let $f: \R^d \to \R^n$. Let $q: \R^d \to \R$. Let $g: \R^d \rightarrow \R^n$. Therefore, we have for any arbitrary $x \in \R^d$, $q(x) \in \R$, $f(x) \in \R^n$, and $g(x) \in \R^n$. Let $a \in \R$ be an arbitrary constant.

Then, we have
    \begin{itemize}
        \item $\frac{\d q(x)^a}{\d x} =  a\cdot q(x)^{a-1} \cdot \frac{\d q(x)}{\d x}$
        \item $\frac {\d \|f(x) \|^2_2}{\d t} = 2 \langle f(x) , \frac{\d f(x)}{\d t} \rangle $
        \item $\frac{\d \langle f(x), g(x) \rangle}{\d t} = \langle \frac{\d f(x)}{\d t} , g(x) \rangle + \langle f(x), \frac{\d g(x)}{\d t} \rangle$
        \item $\frac{\d (g(x) \circ f(x))}{\d t} = \frac{\d g(x)}{\d t} \circ f(x) + g(x) \circ \frac{\d f(x)}{\d t}$ (product rule for Hadamard product)
    \end{itemize}
\end{fact}

\begin{fact}[Basic Vector Norm Bounds]\label{fac:vector_norm}
For vectors $u,v,w \in \R^n$, we have  
\begin{itemize}
    \item Part 1. $\langle u,v \rangle \leq \|u\|_2 \cdot \|v\|_2$ (Cauchy-Schwarz inequality)
    \item Part 2. $ \|\diag (u)  \| \leq \|u \|_\infty$
    \item Part 3. $\|u \circ v \|_2 \leq \|u \|_\infty \cdot \| v\|_2$
    \item Part 4. $\|u \|_\infty \leq \|u \|_2 \leq \sqrt{n} \|u \|_\infty $
    \item Part 5. $\|u \|_2 \leq \|u \|_1  \leq \sqrt{n} \|u \|_2$
    \item Part 6. $ \| \exp(u) \|_\infty \leq \exp(\|u \|_\infty) \leq \exp(\|u \|_2)$
    \item Part 7. Let $\alpha$ be a scalar, then $\|\alpha \cdot u\|_2 = |\alpha| \cdot \| u\|_2$
    \item Part 8. $\|u + v \|_2 \leq \| u\|_2 + \| v
    \|_2 $ 
    \item Part 9. $\|u  v^{\top} \| \leq\| u\|_2\| v\|_2 $
    \item Part 10. 
    if $\| u\|_2 ,\|v \|_2 \leq R$, then $\| \exp(u) - \exp(v) \|_2 \leq \exp(R)\| u -v\|_2$
\end{itemize}    
\end{fact}

\begin{fact}[Matrices Norm Basics]\label{fac:matrix_norm}
For any matrices $U,V \in \R^{n \times n}$, given a scalar $\alpha \in \R$ and a vector $v \in \R^{n}$, we have  
\begin{itemize}
    \item Part 1. $\|U^{\top} \| =\|U \| $ 
    \item Part 2. $\|U \| \geq \|V \| - \|U -V\| $
    \item Part 3. $\|U + V\| \leq \|U \| + \|V \|$
    \item Part 4. $\|U \cdot V\| \leq \|U \| \cdot \|V \|$
    \item Part 5. If $U \preceq \alpha \cdot V$, then $ \|U \| \preceq \alpha \cdot \|V \|$
    \item Part 6. $  \| \alpha \cdot U  \| \leq  |\alpha| \|U \|  $
    \item  Part 7. $\|U v\|_2 \leq\|U \|  \cdot \|v\|_2$
    \item Part 8. $\| U U^\top \| \leq \| U \|^2$
\end{itemize}    
\end{fact}

\begin{fact}[Basic algebraic properties]\label{fac:basic_algebraic_properties}
    Let $x$ be an arbitrary element in $\R$.

    Then, we have
    \begin{itemize}
        \item Part 1. $\exp(x^2) \geq 1$.
        \item Part 2. $\exp(x^2) \geq x$.
    \end{itemize}
\end{fact}

\begin{proof}
\ifdefined\isarxiv
    {\bf Proof of Part 1.}

    Consider
    \begin{align*}
        \frac{\d \exp(x^2)}{\d x} = 2x \exp(x^2) = 0.
    \end{align*}

    This implies that 
    \begin{align*}
        x = 0
    \end{align*}
    since 
    \begin{align*}
        \exp(x^2) \neq 0, \forall x \in \R.
    \end{align*}

    Furthermore, since 
    \begin{align*}
        \frac{\d \exp(x^2)}{\d x} < 0, \text{ when } x < 0
    \end{align*}
    and
    \begin{align*}
        \frac{\d \exp(x^2)}{\d x} > 0, \text{ when } x > 0,
    \end{align*}
    we have that
    \begin{align*}
        (0, \exp(0))
    \end{align*}
    is the local minimum of $\exp(x^2)$.

    Since $x = 0$ is the only critical point of $\exp(x^2)$ and $\exp(x^2)$ is differentiable over all $x \in \R$, so we have
    \begin{align*}
        \exp(x^2) \geq \exp(0^2) = 1,
    \end{align*}
    which completes the proof of the first part.

    {\bf Proof of Part 2.}

    This strategy of proofing this part is the same as the first part by considering the derivative of $\exp(x^2) - x$ and showing that the local minimum of $\exp(x^2) - x$ is greater than $0$, so we omit the proof here.
\else
The proof details can be found in full version \cite{full}.
\fi
\end{proof}

\begin{fact}\label{fac:psd_rule}
    For any vectors $u,v \in \R^n$, we have
    \begin{itemize}
        \item Part 1. $u u^{\top} \preceq \| u\|_2^2 \cdot I_n $
        \item Part 2. $\diag(u) \preceq \|u\|_2 \cdot I_n$
        \item Part 3. $\diag(u \circ u) \preceq \|u\|_2^2 \cdot I_n$
        \item Part 4. $uv^{\top} + vu^{\top} \preceq uu^
        \top + vv^{\top}$
        \item Part 5. $uv^{\top} + vu^{\top} \succeq -( uu^
        \top + vv^{\top})$
        \item Part 6. $(v \circ u) (v \circ u)^{\top} \preceq \| v\|^2_{\infty} u u^{\top}$
        \item  Part 7. $\diag(u \circ v) \preceq \|u\|_2\|v\|_2 \cdot I_n$
    \end{itemize}
\end{fact}

\section{Gradient}
\label{sec:gradient}

In this section, we compute the first-order derivatives of the functions defined earlier.

\begin{lemma}\label{lem:gradient}

Let $x \in \R^d$ be an arbitrary vector. Let $u_1(x), u_2(x), u(x), f(x), c(x), z(x), v_i(x) \in \R^n$ be defined as in Definition~\ref{def:basic_functions}. Let $\alpha(x), L(x), \beta_i(x) \in \R$ be defined as in Definition~\ref{def:basic_functions}.

Then for each $i \in [d]$, we have
\begin{itemize}
    \item Part 1. 
    $\frac{\d u_1(x)}{\d x_i} =  A_{*,i} $ 
    \item Part 2. 
    $\frac{\d u_2(x)}{\d x_i} = u_2(x) \circ A_{*,i} $
    \item Part 3. 
    $\frac{\d u(x)}{\d x_i} =  v_{i}(x)$
    \item Part 4. 
    $\frac{\d \alpha(x)}{\d x_i} =  \beta_i(x)$
    \item Part 5. 
        $\frac{\d \alpha(x)^{-1}}{\d x_i} =  \alpha(x)^{-2} \cdot \beta_i(x)$
    \item Part 6. 
    $\frac{\d f(x)}{\d x_i} = \alpha(x)^{-1} (I_n -  f(x) \cdot {\bf 1}_n^\top)  \cdot  v_i(x) $
    \item Part 7. 
        $\frac{ \d c(x) }{ \d x_i } = \frac{\d f(x)}{\d x_i} = \alpha(x)^{-1} ( I_n  -  f(x)  \cdot  {\bf 1}_n^\top  ) \cdot v_i(x)$
    \item Part 8. 
        $\frac{ \d L(x) }{ \d x_i } = \alpha(x)^{-1}  c(x)^{\top} \cdot (   I_n   -  f(x) \cdot {\bf 1}_n^\top) \cdot v_i(x)$

    \begin{figure}[!ht]
    \centering
    \includegraphics[width = \linewidth]{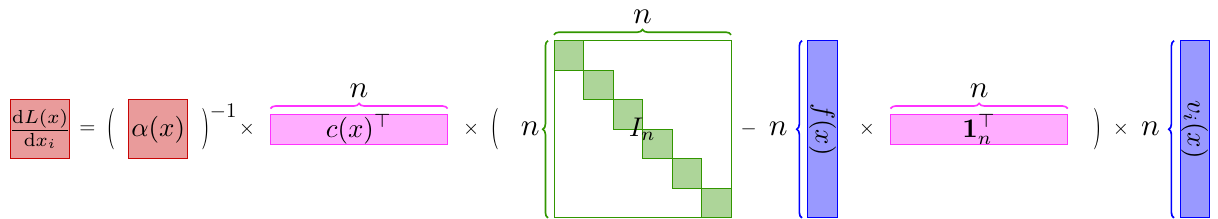}
    \caption{The visualization of Part 8 of Lemma~\ref{lem:gradient}. We have $\alpha(x) \in \R$, $c(x), f(x), {\bf 1}_n, v_i(x) \in \R^n$, and $I_n \in \R^{n \times n}$. First, we subtract the product of $f(x)$ and ${\bf 1}_n^\top$ from $I_n$. Then, we multiply the multiplicative inverse of $\alpha(x)$, $c(x)^\top$, the result from the first step, and $v_i(x)$, which gives us a scalar. The blue rectangles represent the $n$-dimensional vectors. The pink rectangles represent the transposes of $n$-dimensional vectors. The red squares represent the scalar. The green squares represent the identity matrix $I_n$.}
    \label{fig:part_8}
\end{figure}

    \item Part 9. 
        $\frac{\d \beta_i(x)}{\d x_i} = \langle  u_2(x),   A_{*,i} \circ A_{*,i} \rangle$
    \item Part 10. For $j  \in [d] \backslash \{i\}$,
        $\frac{\d \beta_i(x)}{\d x_j} = \langle u_2(x),  A_{*,i} \circ A_{*,j}\rangle$
    \item Part 11. 
            $\frac{\d v_i(x)}{\d x_i} = u_2(x) \circ   A_{*,i} \circ  A_{*,i} $
    \item Part 12. For $j \in [d] \backslash \{ i \}$,
            $\frac{\d v_i(x)}{\d x_j} = u_2(x) \circ  A_{*,j} \circ A_{*,i}$
\end{itemize}
\end{lemma}

\begin{proof}
\ifdefined\isarxiv

{\bf Proof of Part 1.}
For each $i \in [d]$, we have 
\begin{align*}
\frac{\d Ax}{\d x_i} 
 = ~&\frac{A\d x}{\d x_i}\\
 = ~& A_{*,i}
\end{align*}
where the first step follows from simple algebra and the last step follows from the fact that only the $i$-th entry of $\frac{\d x}{\d x_i}$ is $1$ and other entries of it are $0$.

Note that by definition~\ref{def:basic_functions},
\begin{align*}
    u_1(x) = Ax.
\end{align*}

Therefore, we have 
\begin{align*}
     \frac{\d u_1(x)}{\d x_i} =  A_{*,i}.
\end{align*}

{\bf Proof of Part 2.}
For each $i \in [d]$, we have 
\begin{align*}
\frac{\d(u_2(x))_i}{\d x_i} 
= & ~ \frac{\d(\exp(Ax))_i}{\d x_i} \\
= & ~ \exp(Ax)_i \cdot \frac{\d(Ax)_i}{\d x_i} \\
= & ~ \exp(Ax)_i \cdot A_{*,i} \\
= & ~ (u_2(x))_i \cdot A_{*,i},
\end{align*}
where the first step follows from the definition of $(u_2(x))_i$ (see Definition~\ref{def:basic_functions}), the second step follows from Fact~\ref{fac:exp_differention1}, the third step follows from {\bf Part 1}, and the last step follows from the definition of $(u_2(x))_i$ (see Definition~\ref{def:basic_functions}).

Thus, we have  
\begin{align*}
     \frac{\d u_2(x)}{\d x_i} = u_2(x) \circ A_{*,i}
\end{align*}

{\bf Proof of Part 3.}

We have
\begin{align*}
    \frac{\d u(x)}{\d x_i}
    = & ~ \frac{\d (u_1(x)+ u_2(x))}{\d x_i} \\
    = & ~ \frac{\d (u_1(x)) }{\d x_i} + \frac{\d (u_2(x))}{\d x_i} \\
    = & ~ A_{*,i} + u_2(x)  \circ A_{*,i} \\
    = & ~ (u_2(x) + {\bf 1}_n) \circ A_{*,i}  \\
    = & ~ v_i(x),
\end{align*}
where the first step follows from the definition of $u(x)$ (see Definition~\ref{def:basic_functions}), the second step follows from the basic derivative rule, the third step follows from results from {\bf Part 1} and {\bf Part 2}, the fourth step follows from the basic properties of Hadamard product, and the last step follows from the definition of $v_i(x)$ (see Definition~\ref{def:basic_functions}).

{\bf Proof of Part 4.}
\begin{align*}
    \frac{\d \alpha(x)}{\d x_i} 
    = & ~  \frac{\d (\langle u(x) , {\bf 1}_n \rangle)}{\d x_i} \\
    = & ~  \langle \frac{\d u(x)}{\d x_i} , {\bf 1}_n \rangle  \\
    =  &~  \langle v_i(x),  {\bf 1}_n  \rangle\\
    =  & ~ \beta_i(x) 
\end{align*}
where the first step follows from the definition of $\alpha(x)$ (see Definition~\ref{def:basic_functions}), the second step follows from Fact~\ref{fac:exponential_der_rule}, the third step follows from {\bf Part 3}, and the fourth step follows from the definition of $\beta_i(x)$ (see Definition~\ref{def:basic_functions}).

{\bf Proof of Part 5.}
\begin{align*}
    \frac{\d \alpha(x)^{-1}}{\d x_i} 
    = & ~ -1 \cdot \alpha(x)^{-2} \cdot \frac{\d \alpha(x) }{\d x_i}  \\
    = & ~ - \alpha(x)^{-2} \cdot \beta_i(x)
\end{align*}
where the first step follows from the Fact~\ref{fac:exponential_der_rule}, where the second step follows from the results of {\bf Part 4}.

{\bf Proof of Part 6.}

\begin{align*}
    \frac{\d f(x)}{\d x_i} 
    = & ~\frac{\d \alpha(x)^{-1} }{\d x_i} u(x) + \alpha(x)^{-1} \cdot \frac{\d u(x)}{\d x_i} \\
    = & ~- \alpha(x)^{-2} \cdot \beta_i(x) \cdot u(x) + \alpha(x)^{-1} \cdot v_i(x) \\
    = & ~ - \alpha(x)^{-1} f(x) \cdot \beta_i(x) + \alpha(x)^{-1}  \cdot v_i(x) \\
    = & ~\alpha(x)^{-1} \cdot (v_i(x) - f(x) \cdot \beta_i(x))  \\
    = & ~\alpha(x)^{-1} \cdot (v_i(x) - f(x) \cdot \langle v_i(x), {\bf 1}_n \rangle) \\
    = & ~\alpha(x)^{-1} \cdot (v_i(x)- f(x) \cdot {\bf 1}_n^{\top} v_i(x)) \\
    = & ~ \alpha(x)^{-1} \cdot (I_{n} - f(x) \cdot {\bf 1}_n^{\top}) \cdot v_i(x)
 \end{align*}
where the first step follows from the product rule and the definition of $f(x)$ (see Definition~\ref{def:basic_functions}), the second step follows from results of {\bf Part 3, 5}, the third step follows from the definition of $f(x)$ (see Definition~\ref{def:basic_functions}), the fourth step follows from simple algebra, the fifth step follows from the definition of $\beta_i$ (see Definition~\ref{def:basic_functions}), the sixth step follows from Fact~\ref{fac:circ_rules}, and the last step follows from simple algebra.

{\bf Proof of Part 7.}

\begin{align*}
    \frac{ \d c(x) }{ \d x_i } 
    = & ~\frac{ \d (f(x) -b)}{ \d x_i } \\
    = & ~\frac{ \d f(x)}{ \d x_i }  
\end{align*}
where the first step follows from the definition of $c(x)$ (see Definition~\ref{def:basic_functions}), the second step follows from derivative rules.

{\bf Proof of Part 8.}

\begin{align*}
    \frac{\d L(x)}{\d x_i}
    = & ~ \frac{\d 0.5 \| c(x) \|_2^2}{\d x_i}  \\ 
    = & ~ c(x)^{\top} \cdot \frac{\d c(x)}{\d x_i} \\
    = & ~ \alpha(x)^{-1} \cdot c(x)^\top \cdot  (I_{n} - f(x) \cdot {\bf 1}_n^{\top}) \cdot v_i(x)
\end{align*}
where the first step follows from the definition of $L(x)$ (see Definition~\ref{def:basic_functions}), the second step follows from Fact~\ref{fac:exponential_der_rule}, and the last step follows from the results from {\bf Part 6 and 7}. 

{\bf Proof of Part 9.}

\begin{align*}
    \frac{\d \beta_i(x)}{\d x_i} 
    = & ~ \frac{\d (\langle v_i(x) , {\bf 1}_n \rangle)}{\d x_i}\\ 
    = & ~ \frac{\d (\langle (u_2(x) + {\bf 1}_n) \circ A_{*,i} , {\bf 1}_n \rangle)}{\d x_i}\\ 
    = & ~ \frac{\d \langle u_2(x) + {\bf 1}_{n}, A_{*,i} \rangle}{\d x_i}\\
    = & ~ \langle \frac{\d (u_2(x) +{\bf 1}_{n})}{\d x_i}, A_{*,i} \rangle \\
    = & ~ \langle u_2(x) \circ A_{*,i},  A_{*,i}\rangle \\
    = & ~ \langle u_2(x),  A_{*,i} \circ A_{*,i} \rangle 
\end{align*}
where the first step follows from the definition of $\beta_i(x)$ (see Definition~\ref{def:basic_functions}), the second step follows from the definition of $v_i(x)$ (see Definition~\ref{def:basic_functions}), the third step follows from Fact~\ref{fac:circ_rules}, the fourth step follows from Fact~\ref{fac:exponential_der_rule}, the fifth step follows from {\bf Part 2}, and the last step follows from Fact~\ref{fac:circ_rules}.

{\bf Proof of Part 10.}

\begin{align*}
    \frac{\d \beta_i(x)}{\d x_j} 
    = & ~ \frac{\d (\langle v_i(x) , {\bf 1}_n \rangle)}{\d x_j}\\ 
    = & ~ \frac{\d (\langle (u_2(x) + {\bf 1}_n) \circ A_{*,i} , {\bf 1}_n \rangle)}{\d x_j}\\ 
    = & ~ \frac{\d \langle u_2(x) + {\bf 1}_{n}, A_{*,i} \rangle}{\d x_j}\\
    = & ~ \langle \frac{\d (u_2(x) +{\bf 1}_{n})}{\d x_j}, A_{*,i} \rangle \\
    = & ~ \langle u_2(x) \circ A_{*,j},  A_{*,i}\rangle \\
    = & ~ \langle u_2(x),  A_{*,j} \circ A_{*,i} \rangle 
\end{align*}
where the first step follows from the definition of $\beta_i(x)$ (see Definition~\ref{def:basic_functions}), the second step follows from the definition of $v_i(x)$ (see Definition~\ref{def:basic_functions}), the third step follows from Fact~\ref{fac:circ_rules}, the fourth step follows from Fact~\ref{fac:exponential_der_rule}, the fifth step follows from {\bf Part 2}, and the last step follows from Fact~\ref{fac:circ_rules}.

{\bf Proof of Part 11.}
\begin{align*}
    \frac{\d v_i(x)}{\d x_i} 
    = & ~ \frac{\d  (u_2(x) + {\bf 1}_{n}) \circ A_{*,i} }{\d x_i}\\
    = & ~  \frac{\d (u_2(x) +{\bf 1}_{n})}{\d x_i} \circ A_{*,i} \\
    = & ~  u_2(x) \circ A_{*,i} \circ A_{*,i} 
\end{align*}
where the first step follows from the definition of $v_i(x)$ (see Definition~\ref{def:basic_functions}), the second step follows from Fact~\ref{fac:exponential_der_rule} as $\frac{\d A_{*,i}}{\d x_i} = 0$, and the last step follows from the results of {\bf Part 2}.

{\bf Proof of Part 12.}
\begin{align*}
    \frac{\d v_i(x)}{\d x_j} 
    = & ~ \frac{\d  (u_2(x) + {\bf 1}_{n}) \circ A_{*,i} }{\d x_j}\\
    = & ~  \frac{\d (u_2(x) +{\bf 1}_{n})}{\d x_j} \circ A_{*,i} \\
    = & ~ u_2(x) \circ A_{*,j} \circ A_{*,i} 
\end{align*}
where the first step follows from the definition of $v_i(x)$ (see Definition~\ref{def:basic_functions}), the second step follows from Fact~\ref{fac:exponential_der_rule} as $\frac{\d A_{*,i}}{\d x_j} = 0$, and the last step follows from the results of {\bf Part 2}.

\else
The proof details can be found in full version \cite{full}.
\fi
\end{proof}

\section{Hessian}
\label{sec:hessian}

In Section~\ref{sub:hessian:basic_definition}, we introduce the basic definition of the matrices containing $B_1(x), B_2(x), B_3(x) \in \R^{n \times n}$, used for simplifying the expression of Hessian. In Section~\ref{sub:hessian:computation}, we compute the second-order derivatives of the functions defined earlier. In Section~\ref{sub:hessian:lem}, we present a helpful lemma. In Section~\ref{sub:hessian:decompose}, we decompose the matrices $B_1(x), B_2(x), B_3(x) \in \R^{n \times n}$ into low-rank matrices and diagonal matrices.

\subsection{Basic Definition}
\label{sub:hessian:basic_definition}

In this section, we give the definition of the matrices containing $B_1(x), B_2(x), B_3(x) \in \R^{n \times n}$.

\begin{definition}\label{def:B1}
    Let $x \in \R^d$ be an arbitrary vector. Let $u_1(x), u_2(x), u(x), f(x), c(x), z(x), v_i(x) \in \R^n$ and $\alpha(x), L(x), \beta_i(x) \in \R$ be defined as in Definition~\ref{def:basic_functions}. Let $K(x) = (I_n - f(x) \cdot {\bf 1}_n^\top) \in \R^{n \times n}$. Let $\wt{c}(x) = K(x)^\top c(x) \in \R^n$. We define 
\begin{itemize}
    \item $B_1(x) \in \R^{n \times n}$ as
    \begin{align*}
        & ~ 
        A_{*,i}^{\top}B_1(x)A_{*,j}  \\:= & ~ \underbrace{\alpha(x)^{-2}}_{\mathrm{scalar}} \underbrace{v_i(x)^\top}_{1 \times n} \underbrace{K(x)^\top}_{n \times n} \underbrace{K(x)}_{n \times n} \underbrace{v_i(x)}_{n \times 1}
    \end{align*}
    \item $B_2(x) \in \R^{n \times n}$ as
    \begin{align*}
        & ~ A_{*,i}^{\top}B_2(x)A_{*,j} \\
        := & ~ -\underbrace{\alpha(x)^{-2}}_{\mathrm{scalar}}  \cdot \underbrace{\wt{c}(x)^{\top}}_{1 \times n} \cdot (   \underbrace{v_j(x)}_{n \times 1}\cdot \underbrace{\beta_i(x)}_{\mathrm{scalar}} + \underbrace{v_i(x)\cdot \beta_j(x)}_{n \times 1} )
    \end{align*}
    \item $B_3(x) \in \R^{n \times n}$ as
    \begin{align*}
        A_{*,i}^{\top}B_3(x)A_{*,j} := \underbrace{\alpha(x)^{-1}}_{\mathrm{scalar}} \cdot\underbrace{A_{*,i}^\top}_{1 \times n} \mathrm{diag} ( \underbrace{\wt{c}(x)}_{n \times 1} \circ \underbrace{u_2(x)}_{ n \times 1} )   \underbrace{A_{*,j}}_{n \times 1} 
    \end{align*}
\end{itemize}
\end{definition}

\subsection{Computation of Hessian}
\label{sub:hessian:computation}

In this section, we present the computation of Hessian.

\begin{lemma}\label{lem:hessian}
    Let $x \in \R^d$ be an arbitrary vector. Let $u_1(x), u_2(x), u(x), f(x), c(x), z(x), v_i(x) \in \R^n$ and $\alpha(x), L(x), \beta_i(x) \in \R$ be defined as in Definition~\ref{def:basic_functions}. Let $K(x) \in \R^{n \times n}$ and $\wt{c}(x) \in \R^n$ be defined as in Definition~\ref{def:B1}.

    Then for each $i, j \in [d]$, and $j \neq i$, we have
    
    \ifdefined\isarxiv

\begin{itemize}

    \item Part 1. \begin{align*}
        \frac{\d^2 u_1(x)}{\d x_i^2} = {\bf 0}_n
    \end{align*}
    \item Part 2. \begin{align*}
        \frac{\d^2 u_1(x)}{\d x_i \d x_j} = {\bf 0}_n
    \end{align*}
    \item Part 3. \begin{align*}
        \frac{\d^2 u_2(x)}{\d x_i^2} = A_{*,i} \circ u_2(x) \circ A_{*,i}
    \end{align*}
    \item Part 4. \begin{align*}
        \frac{\d^2 u_2(x)}{\d x_i \d x_j} = A_{*,i} \circ u_2(x) \circ A_{*,j}
    \end{align*}

    \item Part 5. \begin{align*}
        \frac{\d^2 u(x)}{\d x_i^2} = A_{*,i} \circ u_2(x) \circ A_{*,i}
    \end{align*}
    \item Part 6. \begin{align*}
        \frac{\d^2 u(x)}{\d x_i \d x_j} = A_{*,i} \circ u_2(x) \circ A_{*,j}
    \end{align*} 

    \item Part 7. \begin{align*}
        \frac{\d^2 \alpha(x)}{\d x_i^2} =\langle u_2(x),  A_{*,i} \circ A_{*,i}  \rangle 
    \end{align*}
    \item Part 8. \begin{align*}
        \frac{\d^2 \alpha(x)}{\d x_i \d x_j} = \langle u_2(x),  A_{*,j} \circ A_{*,i}  \rangle 
    \end{align*}

    \item Part 9. \begin{align*}
        \frac{\d^2 \alpha(x)^{-1}}{\d x_i^2} = 
        \underbrace{\alpha(x)^{-2}}_{\mathrm{scalar}} (\underbrace{\langle  u_2(x) ,  A_{*,i} \circ  A_{*,i}\rangle}_{\mathrm{scalar}} -2 \underbrace{\alpha(x)^{-1} \cdot\beta_i(x)^2)}_{\mathrm{scalar}}
    \end{align*}
    \item Part 10. \begin{align*}
        \frac{\d^2 \alpha(x)^{-1}}{\d x_i \d x_j} = \underbrace{\alpha(x)^{-2}}_{\mathrm{scalar}} (\underbrace{\langle  u_2(x) ,  A_{*,i} \circ  A_{*,j}\rangle}_{\mathrm{scalar}} -2 \underbrace{\alpha(x)^{-1} \cdot \beta_i(x)\beta_j(x))}_{\mathrm{scalar}}
    \end{align*}
    \item Part 11. \begin{align*}
        \frac{\d^2 f(x)}{\d x_i^2} 
        = & ~-2 \underbrace{\alpha(x)^{-2}}_{\mathrm{scalar}} \cdot \underbrace{\beta_i(x)}_{\mathrm{scalar}} \cdot \underbrace{(I_{n} - f(x) \cdot {\bf 1}_n^{\top})}_{n \times n} \cdot \underbrace{v_i(x)}_{n \times 1} + \underbrace{\alpha(x)^{-1}}_{\mathrm{scalar}} \cdot \underbrace{(I_{n} - f(x) \cdot {\bf 1}_n^{\top} )}_{n \times n} \cdot \underbrace{(u_2(x) \circ A_{*,i} \circ A_{*,i})}_{n \times 1}
    \end{align*}
    \item Part 12. \begin{align*}
        \frac{\d^2 f(x)}{\d x_i \d x_j}
        = & ~- \underbrace{\alpha(x)^{-2}}_{\mathrm{scalar}}  \cdot \underbrace{(I_{n} - f(x) \cdot {\bf 1}_n^{\top})}_{n \times n} \cdot( \underbrace{v_j(x)}_{n \times 1}\cdot \underbrace{\beta_i(x)}_{\mathrm{scalar}} + \underbrace{v_i(x)}_{n \times 1}\cdot \underbrace{\beta_j(x)}_{\mathrm{scalar}}) \\
        + & ~ \underbrace{\alpha(x)^{-1}}_{\mathrm{scalar}} \underbrace{(I_{n}- f(x) \cdot {\bf 1}_n^{\top} )}_{n \times n} \cdot \underbrace{(u_2(x) \circ A_{*,j} \circ A_{*,i})}_{n \times 1}
    \end{align*}
    \item Part 13. \begin{align*}
        \frac{\d^2 L(x)}{\d x_i^2} 
        = & ~ \underbrace{\alpha(x)^{-2}}_{\mathrm{scalar}} \cdot \underbrace{A_{*,i}^{\top}}_{1 \times n}\underbrace{B_1(x)}_{n \times n}\underbrace{A_{*,i}}_{n \times 1}- \underbrace{\alpha(x)^{-2}}_{\mathrm{scalar}} \cdot \underbrace{A_{*,i}^{\top}}_{1 \times n}\underbrace{B_2(x)}_{n \times n}\underbrace{A_{*,i}}_{n \times 1}+ \underbrace{\alpha(x)^{-1}}_{\mathrm{scalar}} \cdot \underbrace{A_{*,i}^{\top}}_{1 \times n}\underbrace{B_3(x)}_{n \times n}\underbrace{A_{*,i}}_{n \times 1}
    \end{align*}
    \item Part 14. 
    \begin{align*}
        \frac{\d^2 L(x)}{\d x_i \d x_j}
        = & ~ \underbrace{\alpha(x)^{-2}}_{\mathrm{scalar}} \cdot \underbrace{A_{*,i}^{\top}}_{1 \times n}\underbrace{B_1(x)}_{n \times n}\underbrace{A_{*,j}}_{n \times 1}- \underbrace{\alpha(x)^{-2}}_{\mathrm{scalar}} \cdot \underbrace{A_{*,i}^{\top}}_{1 \times n}\underbrace{B_2(x)}_{n \times n}\underbrace{A_{*,j}}_{n \times 1}+ \underbrace{\alpha(x)^{-1}}_{\mathrm{scalar}} \cdot \underbrace{A_{*,i}^{\top}}_{1 \times n}\underbrace{B_3(x)}_{n \times n}\underbrace{A_{*,j}}_{n \times 1}
    \end{align*}
\end{itemize}

\begin{figure}[!ht]
    \centering
    \includegraphics[width = \linewidth]{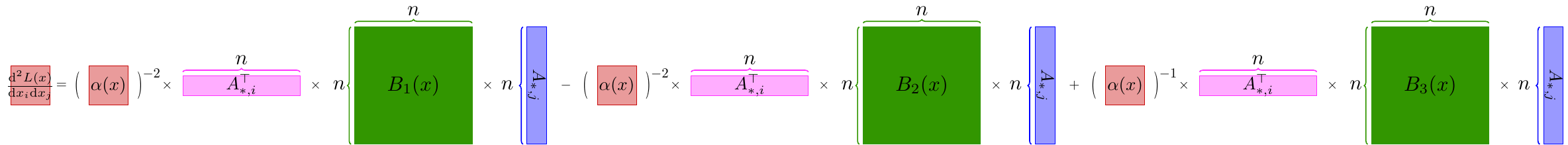}
    \caption{The visualization of Part 14 of Lemma~\ref{lem:hessian}. We have $\alpha(x) \in \R$, $A_{*, i}, A_{*, j} \in \R^n$, and $B_1(x), B_2(x), B_3(x) \in \R^{n \times n}$. First, we multiply $A_{*, i}^\top$ with $B_1(x)$, $B_2(x)$, $B_3(x)$, respectively, and use their product to multiply with $A_{*, j}$. Then, we use $\alpha(x)^{-2}$ to multiply with $A_{*, i}^\top B_1(x) A_{*, j}$ and with $A_{*, i}^\top B_2(x) A_{*, j}$, and we use $\alpha(x)^{-1}$ to multiply with $A_{*, i}^\top B_3(x) A_{*, j}$. Finally, we subtract $\alpha(x)^{-2} A_{*, i}^\top B_2(x) A_{*, j}$ from $\alpha(x)^{-2} A_{*, i}^\top B_1(x) A_{*, j}$, and we add this difference with $\alpha(x)^{-1} A_{*, i}^\top B_3(x) A_{*, j}$. The red squares represent scalars, the pink rectangles represent $1 \times n$ matrices, the green squares represent the $n \times n$ squares, and the blue rectangles represent $n$-dimensional vectors.}
    \label{fig:part_14}
\end{figure}

    \else
    \item Part 1. 
    \begin{align*}
        \frac{\d^2 f(x)}{\d^2 x_i} 
        = & ~-2 \underbrace{\alpha(x)^{-2}}_{\mathrm{scalar}} \cdot \underbrace{\beta_i(x)}_{\mathrm{scalar}} \cdot \underbrace{(I_{n} - f(x) \cdot {\bf 1}_n^{\top})}_{n \times n} \cdot \underbrace{v_i(x)}_{n \times 1} \\
        + & ~ \underbrace{\alpha(x)^{-1}}_{\mathrm{scalar}} \cdot \underbrace{(I_{n} - f(x) \cdot {\bf 1}_n^{\top} )}_{n \times n} \cdot \underbrace{(u_2(x) \circ A_{*,i} \circ A_{*,i})}_{n \times 1}.
    \end{align*}
    \item Part 2. 
    \begin{align*}
        \frac{\d^2 f(x)}{\d x_i \d x_j}
        = & ~ - \underbrace{\alpha(x)^{-2}}_{\mathrm{scalar}}  \cdot \underbrace{(I_{n} - f(x) \cdot {\bf 1}_n^{\top})}_{n \times n} \cdot( \underbrace{v_j(x)}_{n \times 1}\cdot \underbrace{\beta_i(x)}_{\mathrm{scalar}}\\ 
        + & ~ \underbrace{v_i(x)}_{n \times 1}\cdot \underbrace{\beta_j(x)}_{\mathrm{scalar}}) \\
        + & ~ \underbrace{\alpha(x)^{-1}}_{\mathrm{scalar}} \underbrace{(I_{n}- f(x) \cdot {\bf 1}_n^{\top} )}_{n \times n} \cdot \underbrace{(u_2(x) \circ A_{*,j} \circ A_{*,i})}_{n \times 1}
    \end{align*}
    \item Part 3. 
    \begin{align*}
        \frac{\d^2 L(x)}{\d^2 x_i} 
        = & ~ \underbrace{\alpha(x)^{-2}}_{\mathrm{scalar}} \cdot \underbrace{A_{*,i}^{\top}}_{1 \times n}\underbrace{B_1(x)}_{n \times n}\underbrace{A_{*,i}}_{n \times 1}\\
        - & ~ \underbrace{\alpha(x)^{-2}}_{\mathrm{scalar}} \cdot \underbrace{A_{*,i}^{\top}}_{1 \times n}\underbrace{B_2(x)}_{n \times n}\underbrace{A_{*,i}}_{n \times 1}\\
        + & ~ \underbrace{\alpha(x)^{-1}}_{\mathrm{scalar}} \cdot \underbrace{A_{*,i}^{\top}}_{1 \times n}\underbrace{B_3(x)}_{n \times n}\underbrace{A_{*,i}}_{n \times 1}
    \end{align*}
    \item Part 4. 
    \begin{align*}
        \frac{\d^2 L(x)}{\d x_i \d x_j}
        = & ~ \underbrace{\alpha(x)^{-2}}_{\mathrm{scalar}} \cdot \underbrace{A_{*,i}^{\top}}_{1 \times n}\underbrace{B_1(x)}_{n \times n}\underbrace{A_{*,j}}_{n \times 1}\\
        - & ~ \underbrace{\alpha(x)^{-2}}_{\mathrm{scalar}} \cdot \underbrace{A_{*,i}^{\top}}_{1 \times n}\underbrace{B_2(x)}_{n \times n}\underbrace{A_{*,j}}_{n \times 1}\\
        + & ~ \underbrace{\alpha(x)^{-1}}_{\mathrm{scalar}} \cdot \underbrace{A_{*,i}^{\top}}_{1 \times n}\underbrace{B_3(x)}_{n \times n}\underbrace{A_{*,j}}_{n \times 1}
    \end{align*}
    \fi
\end{lemma}

\begin{proof}
\ifdefined\isarxiv

    {\bf Proof of Part 1}
    \begin{align*}
        \frac{\d^2 u_1(x)}{\d x_i^2} 
        = & ~ \frac{\d}{\d x_i} (\frac{\d u_1(x)}{\d x_i})\\
        = & ~ \frac{\d A_{*,i}}{\d x_i}  \\
        = & ~ {\bf 0}_n
    \end{align*}
    where the first step follows from the expansion of the Hessian, the second step follows from {\bf Part 1} of Lemma~\ref{lem:gradient}, and the last step follows from derivative rules.

    {\bf Proof of Part 2}
    \begin{align*}
        \frac{\d^2 u_1(x)}{\d x_i \d x_j}
        =  & ~ \frac{\d}{\d x_j} (\frac{\d u_1(x)}{\d x_i}) \\
        =  & ~ \frac{\d A_{*,i}}{\d x_j} \\
        =  & ~ {\bf 0}_n
    \end{align*}
    where the first step follows from the expansion of the Hessian, the second step follows from {\bf Part 1} of Lemma~\ref{lem:gradient}, and the last step follows from derivative rules.

    {\bf Proof of Part 3}
    \begin{align*}
         \frac{\d^2 u_2(x)}{\d x_i^2} 
         = & ~  \frac{\d}{\d x_i} (\frac{\d u_2(x)}{\d x_i}) \\
         = & ~ \frac{\d (u_2(x) \circ A_{*,i})}{\d x_i} \\
         = & ~ A_{*,i} \circ \frac{\d u_2(x)}{\d x_i} \\
         = & ~ A_{*,i} \circ u_2(x) \circ A_{*,i}
    \end{align*}
    where the first step follows from the expansion of Hessian, the second step follows from {\bf Part 2} of Lemma~\ref{lem:gradient}, the third step follows from Fact~\ref{fac:exponential_der_rule}, and the last step follows from {\bf Part 2} of Lemma~\ref{lem:gradient}.

    {\bf Proof of Part 4}
    \begin{align*}
        \frac{\d^2 u_2(x)}{\d x_ix_j} 
         = & ~  \frac{\d}{\d x_j} (\frac{\d u_2(x)}{\d x_i}) \\
         = & ~ \frac{\d (u_2(x) \circ A_{*,i})}{\d x_j} \\
         = & ~ A_{*,i} \circ \frac{\d u_2(x)}{\d x_j} \\
         = & ~ A_{*,i} \circ u_2(x) \circ A_{*,j}
    \end{align*}
    where the first step follows from the expansion of Hessian, the second step follows from {\bf Part 2} of Lemma~\ref{lem:gradient}, the third step follows from Fact~\ref{fac:exponential_der_rule}, and the last step follows from {\bf Part 2} of Lemma~\ref{lem:gradient}.

    {\bf Proof of Part 5}
    \begin{align*}
         \frac{\d^2 u(x)}{\d x_i^2} 
         = & ~  \frac{\d}{\d x_i} (\frac{\d u_1(x) + u_2(x)}{\d x_i}) \\
         = & ~ \frac{\d}{\d x_i} \frac{\d u_1(x)}{\d x_i} + \frac{\d}{\d x_i}\frac{\d u_2(x)}{\d x_i}\\
         = & ~ A_{*,i} \circ u_2(x) \circ A_{*,i}
        \end{align*}
    where the first step follows from the expansion of Hessian and Definition~\ref{def:basic_functions}, the second step follows from the expansion of derivative, the third step follows from {\bf Part 1} and {\bf Part 3} of this Lemma.

    {\bf Proof of Part 6}
    \begin{align*}
         \frac{\d^2 u(x)}{\d x_ix_j} 
         = & ~  \frac{\d}{\d x_j} (\frac{\d u_1(x) + u_2(x)}{\d x_i}) \\
         = & ~ \frac{\d}{\d x_j} \frac{\d u_1(x)}{\d x_i} + \frac{\d}{\d x_j}\frac{\d u_2(x)}{\d x_i}\\
         = & ~ A_{*,i} \circ u_2(x) \circ A_{*,j}
    \end{align*}
    where the first step follows from the expansion of Hessian and Definition~\ref{def:basic_functions}, the second step follows from the expansion of derivative, the third step follows from {\bf Part 2} and {\bf Part 4} of this Lemma.

    {\bf Proof of Part 7}
    \begin{align*}
        \frac{\d^2 \alpha(x)}{\d x_i^2}
         = & ~  \frac{\d}{\d x_i} (\frac{\d \alpha(x)}{\d x_i}) \\
         = & ~ \frac{\d \beta_i(x)}{\d x_i} \\
         = & ~\langle u_2(x),  A_{*,i} \circ A_{*,i}  \rangle 
    \end{align*}
    where the first step follows from the expansion of Hessian, the second step follows from {\bf Part 4} of Lemma~\ref{lem:gradient}, and the last step follows from {\bf Part 9} of Lemma~\ref{lem:gradient}.

{\bf Proof of Part 8}
    \begin{align*}
        \frac{\d^2 \alpha(x)}{\d x_ix_j}
         = & ~  \frac{\d}{\d x_j} (\frac{\d \alpha(x)}{\d x_i}) \\
         = & ~ \frac{\d \beta_i(x)}{\d x_j} \\
         = & ~ \langle u_2(x) , A_{*,j} \circ A_{*,i}  \rangle
    \end{align*}
where the first step follows from the expansion of Hessian, the second step follows from {\bf Part 4} of Lemma~\ref{lem:gradient}, and the last step follows from {\bf Part 10} of Lemma~\ref{lem:gradient}.

    {\bf Proof of Part 9}
    \begin{align*}
        \frac{\d^2 \alpha(x)^{-1}}{\d x_i^2}
         = & ~  \frac{\d}{\d x_i} (\frac{\d \alpha(x)^{-1}}{\d x_i}) \\
         = & ~ \frac{\d (\alpha(x)^{-2} \cdot \beta_i(x)) }{\d x_i} \\
         = & ~ \frac{\d \alpha(x)^{-2} }{\d x_i} \cdot \beta_i(x) +\alpha(x)^{-2} \cdot \frac{\d \beta_i(x) }{\d x_i} \\
         = & ~ -2  \alpha(x)^{-3} \cdot \frac{\d\alpha(x)}{\d x_i} \cdot \beta_i(x) +  \alpha(x)^{-2} \cdot \langle  u_2(x) ,  A_{*,i} \circ  A_{*,i}\rangle \\
         = & ~ -2  \alpha(x)^{-3} \cdot \beta_i(x)^2 + \alpha(x)^{-2} \cdot \langle  u_2(x) ,  A_{*,i} \circ  A_{*,i}\rangle \\
         = & ~\alpha(x)^{-2} (\langle  u_2(x) ,  A_{*,i} \circ  A_{*,i}\rangle -2 \alpha(x)^{-1} \cdot \beta_i(x)^2)
         \end{align*} 
where the first step follows from the expansion of Hessian, the second step follows from {\bf Part 5} of Lemma~\ref{lem:gradient}, the third step follows from the product rule of derivative, the fourth step follows from Fact~\ref{fac:exponential_der_rule} and {\bf Part 9} of Lemma~\ref{lem:gradient}, the fifth step follows from {\bf Part 4} of Lemma~\ref{lem:gradient}, and the last step follows from simple algebra.

     {\bf Proof of Part 10}
        \begin{align*}
        \frac{\d^2 \alpha(x)^{-1}}{\d x_ix_j}
         = & ~  \frac{\d}{\d x_j} (\frac{\d \alpha(x)^{-1}}{\d x_i}) \\
         = & ~ \frac{\d (\alpha(x)^{-2} \cdot \beta_i(x)) }{\d x_j} \\
         = & ~ \frac{\d \alpha(x)^{-2} }{\d x_j} \cdot \beta_i(x) +\alpha(x)^{-2} \cdot \frac{\d \beta_i(x) }{\d x_j} \\
         = & ~ -2  \alpha(x)^{-3} \cdot \frac{\d\alpha(x)}{\d x_j} \cdot \beta_i(x) +  \alpha(x)^{-2} \cdot \langle  u_2(x) ,  A_{*,j} \circ  A_{*,i}\rangle \\
         = & ~ -2  \alpha(x)^{-3} \cdot \beta_i(x) \cdot \beta_j(x) + \alpha(x)^{-2} \cdot \langle  u_2(x) ,  A_{*,j} \circ  A_{*,i}\rangle \\
         = & ~\alpha(x)^{-2} (\langle  u_2(x) ,  A_{*,j} \circ  A_{*,i}\rangle -2 \alpha(x)^{-1} \cdot \beta_i(x) \cdot \beta_j(x))
         \end{align*} 
where the first step follows from the expansion of Hessian, the second step follows from {\bf Part 5} of Lemma~\ref{lem:gradient}, the third step follows from the product rule of derivative, the fourth step follows from Fact~\ref{fac:exponential_der_rule} and {\bf Part 10} of Lemma~\ref{lem:gradient}, the fifth step follows from {\bf Part 4} of Lemma~\ref{lem:gradient}, and the last step follows from simple algebra.

    {\bf Proof of Part 11}

We first analyze the following equation:
\begin{align}\label{eq:pre_second_derivative_of_f}
\alpha(x)^{-1} \cdot (\frac{\d (I_{n} - f(x)\cdot {\bf 1}_n^{\top})}{\d x_i}\cdot  v_i(x) )
= & ~ \alpha(x)^{-1} \cdot (\frac{\d I_{n}}{\d x_i} - \frac{\d f(x)\cdot {\bf 1}_n^{\top}}{\d x_i}\cdot  v_i(x) ) \notag\\
= & ~ \alpha(x)^{-1} \cdot ( - \frac{\d f(x)}{\d x_i} \cdot {\bf 1}_n^{\top} \cdot  v_i(x) ) \notag\\
= & ~ \alpha(x)^{-1} \cdot ( - \alpha(x)^{-1} (I_n -  f(x) \cdot {\bf 1}_n^\top)  \cdot  v_i(x) \cdot {\bf 1}_n^{\top} \cdot  v_i(x) ) \notag\\
= & ~ -\alpha(x)^{-2} \cdot (I_n -  f(x) \cdot {\bf 1}_n^\top)  \cdot  \beta_i(x) \cdot  v_i(x),
\end{align}
where the first step follows from the basic derivative rule, the second step follows from the product rule, the third step follows from {\bf Part 6} of Lemma~\ref{lem:gradient}, and the last step follows from simple algebra and the definition of $\beta_i(x)$ (see Definition~\ref{def:basic_functions}).

Then, we have
    \begin{align*}
        \frac{\d^2 f(x)}{\d x_i^2} 
        = & ~  \frac{\d}{\d x_i} (\frac{\d f(x)}{\d x_i}) \\
        = & ~ \frac{\d (\alpha(x)^{-1} \cdot (I_{n} - f(x) \cdot {\bf 1}_n^{\top}) \cdot v_i(x) )}{\d x_i} \\
        = & ~ \frac{\d \alpha(x)^{-1}}{\d x_i} \cdot (I_{n} - f(x) \cdot {\bf 1}_n^{\top}) \cdot v_i(x)  + \alpha(x)^{-1} \cdot \frac{\d (I_{n} - f(x) \cdot {\bf 1}_n^{\top}) \cdot v_i(x)}{\d x_i} \\
        = & ~ - \alpha(x)^{-2} \cdot \beta_i(x) \cdot (I_{n} - f(x) \cdot {\bf 1}_n^{\top}) \cdot v_i(x) \\
        + & ~ \alpha(x)^{-1} \cdot (\frac{\d (I_{n} - f(x)\cdot {\bf 1}_n^{\top})}{\d x_i}\cdot  v_i(x) +(I_{n} - f(x)\cdot {\bf 1}_n^{\top})\cdot \frac{\d v_i(x)}{\d x_i} )\\
        = & ~ - \alpha(x)^{-2} \cdot \beta_i(x) \cdot (I_{n} - f(x) \cdot {\bf 1}_n^{\top}) \cdot v_i(x) \\
        + & ~ -\alpha(x)^{-2} \cdot (I_{n} - f(x) \cdot {\bf 1}_n^{\top} )  \cdot v_i(x) \cdot \beta_i(x)  + \alpha(x)^{-1} \cdot (I_{n} - f(x) \cdot {\bf 1}_n^{\top} ) \cdot (u_2(x) \circ A_{*,i} \circ A_{*,i}) \\
        = & ~-2 \alpha(x)^{-2} \cdot \beta_i(x) \cdot (I_{n} - f(x) \cdot {\bf 1}_n^{\top}) \cdot v_i(x) + \alpha(x)^{-1} \cdot (I_{n} - f(x) \cdot {\bf 1}_n^{\top} ) \cdot (u_2(x) \circ A_{*,i} \circ A_{*,i})
    \end{align*}
where the first step follows from the expansion of Hessian, the second step follows from {\bf Part 6} of Lemma~\ref{lem:gradient}, the third step follows from the product rule of derivative, the fourth step follows from {\bf Part 5} of Lemma~\ref{lem:gradient} and the product rule, the fifth step follows Eq.~\eqref{eq:pre_second_derivative_of_f} and {\bf Part 11} of Lemma~\ref{lem:gradient}, and the last step follows from simple algebra.

{\bf Proof of Part 12}

We first analyze the following equation:
\begin{align}\label{eq:pre_second_derivative_of_fj}
\alpha(x)^{-1} \cdot (\frac{\d (I_{n} - f(x)\cdot {\bf 1}_n^{\top})}{\d x_j}\cdot  v_i(x) )
= & ~ \alpha(x)^{-1} \cdot (\frac{\d I_{n}}{\d x_j} - \frac{\d f(x)\cdot {\bf 1}_n^{\top}}{\d x_j}\cdot  v_i(x) ) \notag\\
= & ~ \alpha(x)^{-1} \cdot ( - \frac{\d f(x)}{\d x_j} \cdot {\bf 1}_n^{\top} \cdot  v_i(x) ) \notag\\
= & ~ \alpha(x)^{-1} \cdot ( - \alpha(x)^{-1} (I_n -  f(x) \cdot {\bf 1}_n^\top)  \cdot  v_j(x) \cdot {\bf 1}_n^{\top} \cdot  v_i(x) ) \notag\\
= & ~ -\alpha(x)^{-2} \cdot (I_n -  f(x) \cdot {\bf 1}_n^\top)  \cdot  \beta_i(x) \cdot  v_j(x),
\end{align}
where the first step follows from the basic derivative rule, the second step follows from the product rule, the third step follows from {\bf Part 6} of Lemma~\ref{lem:gradient}, and the last step follows from simple algebra and the definition of $\beta_i(x)$ (see Definition~\ref{def:basic_functions}).

Then, we have
\begin{align*}
     \frac{\d^2 f(x)}{\d x_ix_j} 
        = & ~  \frac{\d}{\d x_j} (\frac{\d f(x)}{\d x_i}) \\
        = & ~ \frac{\d (\alpha(x)^{-1} \cdot (I_{n} - f(x) \cdot {\bf 1}_n^{\top}) \cdot v_i(x) )}{\d x_j} \\
        = & ~ \frac{\d \alpha(x)^{-1}}{\d x_j} \cdot (I_{n} - f(x) \cdot {\bf 1}_n^{\top}) \cdot v_i(x)  + \alpha(x)^{-1} \cdot \frac{\d (I_{n} - f(x) \cdot {\bf 1}_n^{\top}) \cdot v_i(x))}{\d x_j} \\
        = & ~ - \alpha(x)^{-2} \cdot \beta_j(x) \cdot (I_{n} - f(x) \cdot {\bf 1}_n^{\top}) \cdot v_i(x) \\
        + & ~ \alpha(x)^{-1} \cdot (\frac{\d (I_{n} - f(x)\cdot {\bf 1}_n^{\top})}{\d x_j}\cdot  v_i(x) +(I_{n} - f(x)\cdot {\bf 1}_n^{\top})\cdot \frac{\d v_i(x)}{\d x_j} )\\
        = & ~ - \alpha(x)^{-2} \cdot \beta_j(x) \cdot (I_{n} - f(x) \cdot {\bf 1}_n^{\top}) \cdot v_i(x) \\
        + & ~ -\alpha(x)^{-2} \cdot (I_{n} - f(x) \cdot {\bf 1}_n^{\top} )  \cdot v_j(x) \cdot \beta_i(x)  + \alpha(x)^{-1} \cdot (I_{n} - f(x) \cdot {\bf 1}_n^{\top} ) \cdot (u_2(x) \circ A_{*,j} \circ A_{*,i}) \\
        = & ~- \alpha(x)^{-2}  \cdot (I_{n} - f(x) \cdot {\bf 1}_n^{\top}) \cdot(v_j(x)\cdot \beta_i(x) + v_i(x)\cdot \beta_j(x)) \\
        + & ~ \alpha(x)^{-1} (I_{n}- f(x) \cdot {\bf 1}_n^{\top} ) \cdot (u_2(x) \circ A_{*,j} \circ A_{*,i})
\end{align*}
where the first step follows from the expansion of Hessian, the second step follows from {\bf Part 6} of Lemma~\ref{lem:gradient}, the third step follows from the product rule of derivative, the fourth step follows from {\bf Part 5} of Lemma~\ref{lem:gradient} and product rule, the fifth step follows from Eq.~\eqref{eq:pre_second_derivative_of_fj} and {\bf Part 12} of Lemma~\ref{lem:gradient}, and the last step follows from simple algebra.

    {\bf Proof of Part 13}
    \begin{align*}
        \frac{\d^2 L(x)}{\d x_i^2} 
        = & ~\frac{\d}{\d x_i} (\frac{\d L(x)}{\d x_i}) \\
        = & ~ \frac{\d}{\d  x_i} \langle c(x) , \frac{\d c(x)}{\d x_i} \rangle\\
        = & ~ \frac{\d}{\d  x_i} \langle c(x) , \frac{\d f(x)}{\d x_i} \rangle\\
        = & ~ \langle \frac{\d c(x)}{\d x_i} ,  \frac{\d f(x)}{\d x_i} \rangle + c(x)^{\top} \cdot \frac{\d^2 f(x)}{\d^2 x_i} \\
        = & ~(\alpha(x)^{-1}  \cdot  (I_{n} - f(x) \cdot {\bf 1}_n^{\top}) \cdot v_i(x))^{\top} \cdot \alpha(x)^{-1} \cdot   (I_{n} - f(x) \cdot {\bf 1}_n^{\top}) \cdot v_i(x) \\
        + & ~c(x)^{\top} \cdot -2 \alpha(x)^{-2} \cdot \beta_i(x) \cdot (I_{n} - f(x) \cdot {\bf 1}_n^{\top}) \cdot v_i(x) \\
        + & ~ \alpha(x)^{-1} \cdot c(x)^{\top} \cdot (I_{n} - f(x) \cdot {\bf 1}_n^{\top} ) \cdot (u_2(x) \circ A_{*,i} \circ A_{*,i})\\
        = & ~ \alpha(x)^{-2} v_i(x)^\top K(x)^\top K(x) v_i(x) \\
        + & ~ - 2\alpha(x)^{-2}  \cdot \wt{c}(x)^{\top} \cdot   v_i(x)\cdot \beta_i(x)  \\
        + & ~ \alpha(x)^{-1} \cdot A_{*,i}^\top \mathrm{diag} ( \wt{c}(x) \circ u_2(x) )   A_{*,i} \\
        = & ~  A_{*,i}^{\top}B_1(x)A_{*,i} + A_{*,i}^{\top}B_2(x)A_{*,i} + A_{*,i}^{\top}B_3(x)A_{*,i}
    \end{align*}
where the first step follows from the expansion of Hessian, the second step follows from Fact~\ref{fac:exponential_der_rule}, the third step follows from {\bf Part 7} of Lemma~\ref{lem:gradient}, the fourth step follows from Fact~\ref{fac:exponential_der_rule}, the fifth step follows from {\bf Part 7} of Lemma~\ref{lem:gradient} and {\bf Part 11} of Lemma~\ref{lem:hessian}, the sixth step follows from Definition of $\wt{c}, K$ (See Definition~\ref{def:B1}), and the last step follow from Definitions of $B_1,B_2,B_3$ (See Definition~\ref{def:B1}).

    {\bf Proof of Part 14}
\begin{align*}
        \frac{\d^2 L(x)}{\d x_ix_j} 
        = & ~\frac{\d}{\d x_j} (\frac{\d L(x)}{\d x_i}) \\
        = & ~ \frac{\d}{\d  x_j} \langle c(x) , \frac{\d c(x)}{\d x_i} \rangle\\
        = & ~ \frac{\d}{\d  x_j} \langle c(x) , \frac{\d f(x)}{\d x_i} \rangle\\
        = & ~\frac{\d c(x)^{\top}}{\d x_j}\cdot \frac{\d f(x)}{\d x_i} + c(x)^{\top} \cdot \frac{\d^2 f(x)}{\d x_ix_j} \\
        = & ~(\alpha(x)^{-1}  \cdot  (I_{n} - f(x) \cdot {\bf 1}_n^{\top}) \cdot v_j(x))^{\top} \cdot \alpha(x)^{-1} \cdot   (I_{n} - f(x) \cdot {\bf 1}_n^{\top}) \cdot v_i(x) \\
        + & ~c(x)^{\top} \cdot ( - \alpha(x)^{-2}  \cdot (I_{n} - f(x) \cdot {\bf 1}_n^{\top}) \cdot(v_j(x)\cdot \beta_i(x) + v_i(x)\cdot \beta_j(x))\\
        + & ~ \alpha(x)^{-1} (I_{n}- f(x) \cdot {\bf 1}_n^{\top} ) \cdot (u_2(x) \circ A_{*,j} \circ A_{*,i})) \\
        = & ~ \alpha(x)^{-2} v_i(x)^\top K(x)^\top K(x) v_j(x) \\
        + & ~ -\alpha(x)^{-2}  \cdot \wt{c}(x)^{\top} \cdot (   v_j(x)\cdot \beta_i(x) + v_i(x)\cdot \beta_j(x) ) \\
        + & ~ \alpha(x)^{-1} \cdot A_{*,i}^\top \mathrm{diag} ( \wt{c}(x) \circ u_2(x) )   A_{*,j} \\
        = & ~ A_{*,i}^{\top}B_1(x)A_{*,j} + A_{*,i}^{\top}B_2(x)A_{*,j} + A_{*,i}^{\top}B_3(x)A_{*,j}
    \end{align*}
    where the first step follows from the expansion of Hessian, the second step follows from Fact~\ref{fac:exponential_der_rule}, the third step follows from {\bf Part 7} of Lemma~\ref{lem:gradient}, the fourth step follows from Fact~\ref{fac:exponential_der_rule}, the fifth step follows from {\bf Part 7} of Lemma~\ref{lem:gradient} and {\bf Part 12} of Lemma~\ref{lem:hessian}, the sixth step follows from Definition of $\wt{c}, K$ (See Definition~\ref{def:B1}), and the last step follow from Definitions of $B_1,B_2,B_3$ (See Definition~\ref{def:B1}). 
\else
The proof details can be found in full version \cite{full}.
\fi
\end{proof}

\subsection{Helpful Lemma}
\label{sub:hessian:lem}

In this section, we present a helpful lemma that is used for further analysis of Hessian. 

\begin{lemma}\label{lem:help_lemma}
Let $x \in \R^d$ be an arbitrary vector. Let $u_1(x), u_2(x), u(x), f(x), c(x), z(x), v_i(x) \in \R^n$ and $\alpha(x), L(x), \beta_i(x) \in \R$ be defined as in Definition~\ref{def:basic_functions}. Let $K(x) \in \R^{n \times n}$ and $\wt{c}(x) \in \R^n$ be defined as in Definition~\ref{def:B1}.
    
Then, for each $i,j \in [d]$,
    \begin{itemize}
        \item Part 1. 
        \begin{align*}
        & ~ \alpha(x)^{-2} \cdot v_i(x)^\top K(x)^\top K(x) v_j(x) \\
        = & ~  \underbrace{A_{*,i}^{\top}}_{1 \times n} \cdot \underbrace{\alpha(x)^{-2} }_{\mathrm{scalar}}  \cdot \underbrace{\diag(z(x))}_{n \times n}  \\
        \cdot & ~ \underbrace{K(x)^\top}_{n \times n } \underbrace{K(x)}_{n \times n} \cdot \diag(\underbrace{z(x)}_{n \times 1})\cdot  \underbrace{A_{*,j}}_{n \times 1
        }            
        \end{align*} 
        \item Part 2. 
        \begin{align*}
       & ~ \alpha(x)^{-2} \cdot \wt{c}(x)^{\top} \cdot (   v_j(x)\cdot \beta_i(x) + v_i(x)\cdot \beta_j(x) ) \\
       = & ~  \underbrace{A_{*,i}^{\top}}_{1 \times n} \cdot \underbrace{ \alpha(x)^{-2} }_{ \mathrm{scalar} } \cdot \underbrace{z(x)}_{n \times 1} \cdot  \underbrace{\wt{c}(x)^{\top}}_{1 \times n}  \cdot \underbrace{\diag(z(x))}_{n \times n} \cdot  \underbrace{A_{*,j}}_{n \times 1}\\
        + & ~ \underbrace{A_{*,i}^{\top}}_{1 \times n} \cdot \underbrace{ \alpha(x)^{-2} }_{ \mathrm{scalar} }  \cdot \underbrace{\diag(z(x))}_{n \times n} \cdot \underbrace{\wt{c}(x)}_{n \times 1} \cdot  \underbrace{z(x)^{\top}}_{1 \times n}  \cdot \underbrace{A_{*,j}}_{n \times 1}
       \end{align*}
       \item Part 3.
       \begin{align*}
           & ~ \alpha(x)^{-1} \cdot A_{*,i}^\top \mathrm{diag} ( \wt{c}(x) \circ u_2(x) )   A_{*,j} \\
           = & ~ \underbrace{A_{*,i}^\top}_{1 \times n} \cdot \underbrace{\alpha(x)^{-1}}_{\mathrm{scalar}} \cdot \underbrace{\mathrm{diag} ( \wt{c}(x) \circ u_2(x) )}_{n \times n}   \underbrace{A_{*,j}}_{n \times 1}
       \end{align*}
    \end{itemize}

\end{lemma}

\begin{proof}
\ifdefined\isarxiv

        {\bf Proof of Part 1.}
        \begin{align*}
            & ~ \alpha(x)^{-2} \cdot v_i(x)^\top K(x)^\top K(x) v_j(x) \\
            = & ~  \alpha(x)^{-2} \cdot ((z(x) \circ A_{*,i})^{\top} \cdot  K(x)^\top K(x) \cdot ( z(x) \circ A_{*,j}))\\
            = & ~ \alpha(x)^{-2} \cdot (\diag( z(x) ) \cdot A_{*,i})^{\top} \cdot  K(x)^\top K(x) \cdot \diag( z(x) ) \cdot  A_{*,j}\\
            = & ~  A_{*,i}^{\top} \cdot \alpha(x)^{-2} \cdot \diag(z(x))  \cdot  K(x)^\top K(x) \cdot \diag(z(x))\cdot  A_{*,j}
        \end{align*}
where the first step follows from the definition of $v_i(x)$ (see Definition~\ref{def:basic_functions}), the second step follows from Fact~\ref{fac:circ_rules}, and the last step follows from simple algebra and the definition of $z(x)$ (see Definition~\ref{def:basic_functions}).

        {\bf Proof of Part 2.}
        \begin{align*}
            & ~ \alpha(x)^{-2} \cdot \wt{c}(x)^{\top} \cdot (   v_j(x)\cdot \beta_i(x) + v_i(x)\cdot \beta_j(x) )  \\
            = & ~ \alpha(x)^{-2} \cdot\wt{c}(x)^{\top} \cdot z(x) \circ A_{*,j} \cdot \langle z(x) , A_{*,i} \rangle \\
            + & ~   \alpha(x)^{-2} \cdot\wt{c}(x)^{\top} \cdot  z(x) \circ A_{*,i} \cdot \langle z(x) , A_{*,j} \rangle \\ 
            = & ~  \alpha(x)^{-2} \cdot\wt{c}(x)^{\top} \diag( z(x) ) \cdot A_{*,j} \cdot z(x)^{\top} \cdot A_{*,i} \\
            + & ~ \alpha(x)^{-2} \cdot\wt{c}(x)^{\top} \diag( z(x) ) \cdot A_{*,i} \cdot z(x) ^{\top} \cdot A_{*,j} \\
            = & ~  (A_{*,j}^{\top} \cdot \alpha(x)^{-2} \cdot \diag( z(x) ) \cdot \wt{c}(x) \cdot z(x)^{\top} \cdot A_{*,i})^{\top} \\
            + & ~ A_{*,i}^{\top} \cdot \alpha(x)^{-2} \cdot\diag( z(x) ) \cdot \wt{c}(x) \cdot ( z(x) )^{\top} \cdot A_{*,j} \\
            = & ~  A_{*,i}^{\top} \cdot \alpha(x)^{-2} \cdot z(x) \cdot \wt{c}(x)^{\top} \cdot \diag (z(x))  \cdot A_{*,j} \\
            + & ~ A_{*,i}^{\top} \cdot \alpha(x)^{-2} \cdot \diag (z(x)) \cdot \wt{c}(x) \cdot z(x)^{\top} \cdot A_{*,j}
\end{align*}
where the first step follows from the definition of $\beta_i(x)$ and $v_i(x)$ (see Definition~\ref{def:basic_functions}), the second step follows from Fact~\ref{fac:circ_rules}, the third step follows from Fact~\ref{fac:circ_rules}, and the last step follows from simple algebra and the definition of $z(x)$ (see Definition~\ref{def:basic_functions}).

        {\bf Proof of Part 3.}
        \begin{align*}
            & ~ \alpha(x)^{-1} \cdot A_{*,i}^\top \mathrm{diag} ( \wt{c}(x) \circ u_2(x) )   A_{*,i} \\
            = & ~  A_{*,i}^\top \cdot \alpha(x)^{-1} \cdot \mathrm{diag} ( \wt{c}(x) \circ u_2(x) )   A_{*,i}
        \end{align*}
where the first step follows from the simple algebra.

\else
The proof details can be found in full version \cite{full}.
\fi
\end{proof}

\subsection{Decomposing \texorpdfstring{$B_1(x), B_2(x)$ and $B_3(x)$}{} into low rank plus diagonal} 
\label{sub:hessian:decompose}

In this section, we decompose the matrices $B_1(x)$, $B_2(x)$, and $B_3(x)$ into low rank plus diagonal.

\begin{lemma}\label{lem:decompose}

    Let $x \in \R^d$ be an arbitrary vector. Let $u_1(x), u_2(x), u(x), f(x), c(x), z(x), v_i(x) \in \R^n$ and $\alpha(x), L(x), \beta_i(x) \in \R$ be defined as in Definition~\ref{def:basic_functions}. Let $K(x), B_1(x), B_2(x), B_3(x) \in \R^{n \times n}$ and $\wt{c}(x) \in \R^n$ be defined as in Definition~\ref{def:B1} and $B(x) = B_1(x) + B_2(x) + B_3(x) \in \R^{n \times n}$.

Then, we show that
\begin{itemize}
    \item Part 1. For $B_1(x) \in \R^{n \times n}$, we have 
    \begin{align*}
        B_1(x) = \underbrace{\alpha(x)^{-2} }_{\mathrm{scalar}}  \cdot \underbrace{\diag(z(x))}_{n \times n}  \cdot  \underbrace{K(x)^\top}_{n \times n } \underbrace{K(x)}_{n \times n} \cdot \diag(\underbrace{z(x)}_{n \times 1})
    \end{align*}
    \item Part 2.  For $B_2(x) \in \R^{n \times n}$, we have
    \begin{align*}
         B_2(x) 
         = & ~ -\underbrace{ \alpha(x)^{-2} }_{ \mathrm{scalar} } \cdot \underbrace{\wt{c}(x)}_{n \times 1} \cdot  \underbrace{z(x)^{\top}}_{1 \times n} \cdot \underbrace{\diag(z(x))}_{n \times n}\\  
         - & ~ \underbrace{ \alpha(x)^{-2} }_{ \mathrm{scalar} } \cdot\underbrace{\diag(z(x))}_{n \times n} \cdot \underbrace{z(x)}_{n \times 1} \cdot  \underbrace{\wt{c}(x)^{\top}}_{1 \times n}
    \end{align*}
    \item Part 3.  For $B_3(x) \in \R^{n \times n}$, we have
    \begin{align*}
         B_3(x) = \underbrace{\alpha(x)^{-1}}_{\mathrm{scalar}} \cdot \mathrm{diag} ( \underbrace{\wt{c}(x)}_{n \times 1} \circ \underbrace{u_2(x)}_{n \times 1} ) 
    \end{align*}
    \item Part 4. For $B(x) \in \R^{n \times n}$, we have 
    \begin{align*}
        B(x) 
        = & ~ \underbrace{\alpha(x)^{-2}}_{\mathrm{scalar}} \underbrace{\diag(z(x))}_{n \times n}  \cdot  \underbrace{K(x)^\top}_{n \times n } \underbrace{K(x)}_{n \times n} \cdot \diag(\underbrace{z(x)}_{n \times 1})\\
        - & ~ \underbrace{ \alpha(x)^{-2} }_{ \mathrm{scalar} } \cdot \underbrace{\wt{c}(x)}_{n \times 1} \cdot  \underbrace{z(x)^{\top}}_{1 \times n} \cdot \underbrace{\diag(z(x))}_{n \times n}  \\
        - & ~ \underbrace{ \alpha(x)^{-2} }_{ \mathrm{scalar} } \cdot\underbrace{\diag(z(x))}_{n \times n} \cdot \underbrace{z(x)}_{n \times 1} \cdot  \underbrace{\wt{c}(x)^{\top}}_{1 \times n}\\
        + & ~ \underbrace{\alpha(x)^{-1}}_{\mathrm{scalar}} \cdot \mathrm{diag} ( \underbrace{\wt{c}(x)}_{n \times 1} \circ \underbrace{u_2(x)}_{n \times 1} ) 
    \end{align*}
    
\end{itemize}

\end{lemma}

\begin{proof}
\ifdefined\isarxiv

    {\bf Proof of Part 1}
    \begin{align*}
        A_{*,i}^{\top}B_1(x)A_{*,j} 
        = & ~ \underbrace{\alpha(x)^{-2}}_{\mathrm{scalar}} \cdot\underbrace{v_i(x)^\top}_{1 \times n} \underbrace{K(x)^\top}_{n \times n} \underbrace{K(x)}_{n \times n} \underbrace{v_i(x)}_{n \times 1} \\
        = & ~  A_{*,i}^{\top} \cdot \alpha(x)^{-2} \cdot \diag(z(x))^{\top}  \cdot K(x)^\top K(x) \cdot \diag(z(x))\cdot  A_{*,j}
    \end{align*}
    where the first step follows from Definition~\ref{def:B1}, and the last step follows from Lemma~\ref{lem:help_lemma}.

    Thus, by extracting $A_{*,i}^{\top}$ and $A_{*,j}$, we get:
    \begin{align*}
        B_1(x) = \alpha(x)^{-2} \cdot \diag(z(x))^{\top}  \cdot  K(x)^\top K(x) \cdot \diag(z(x))
    \end{align*}

    {\bf Proof of Part 2.}
    \begin{align*}
        & ~ A_{*,i}^{\top}B_2(x) A_{*,j}\\
        = & ~ -\alpha(x)^{-2} \cdot \wt{c}(x)^{\top} \cdot (   v_j(x)\cdot \beta_i(x) + v_i(x)\cdot \beta_j(x) ) \\
        =  & ~ -(A_{*,i}^{\top} \cdot \alpha(x)^{-2} \cdot z(x) \cdot \wt{c}(x)^{\top} \cdot \diag (z(x))  \cdot A_{*,j}  \\
        + & ~ A_{*,i}^{\top} \cdot \alpha(x)^{-2} \cdot \diag (z(x)) \cdot \wt{c}(x) \cdot z(x)^{\top} \cdot A_{*,j})
    \end{align*}
    where the first step follows from the Definition of $A_{*,i}^{\top}B_2(x) A_{*,j}$ (see Definition~\ref{def:B1}), and the last step follows from Lemma~\ref{lem:help_lemma}.
    
    Thus, by extracting $A_{*,i}^{\top}$ and $A_{*,j}$, we get:
    \begin{align*}
        B_2(x) 
        = & ~ -(\alpha(x)^{-2} \cdot z(x) \cdot \wt{c}(x)^{\top} \cdot \diag (z(x)) \\
        + & ~ \alpha(x)^{-2} \cdot \diag (z(x)) \cdot \wt{c}(x) \cdot z(x)^{\top})
    \end{align*}

    {\bf Proof of Part 3.}
    \begin{align*}
         A_{*,i}^\top B_3(x)A_{*,j}
        = & ~  A_{*,i}^\top \cdot \alpha(x)^{-1} \cdot \mathrm{diag} ( \wt{c}(x) \circ u_2(x) )   A_{*,j} 
    \end{align*}
    where the first step follows from Lemma~\ref{lem:help_lemma}.

    Thus, by extracting $A_{*,i}^{\top}$ and $A_{*,j}$, we get:
    \begin{align*}
        B_3(x) = \alpha(x)^{-1} \cdot \mathrm{diag} ( \wt{c}(x) \circ u_2(x) )
    \end{align*}

    {\bf Proof of Part 4.}

    Since $B(x) = B_1(x) + B_2(x) + B_3(x)$, by combining the first three part, we can get $B(x)$.

\else
The proof details can be found in full version \cite{full}.
\fi
\end{proof}

\section{Rewrite Hessian}
\label{sec:rewrite_hessian}

In this section, we rewrite the hessian. For convenience of analysis, we formally make a definition block for $B(x)$.

\begin{definition}\label{def:B}

Let $x \in \R^d$ be an arbitrary vector. Let $u_1(x), u_2(x), u(x), f(x), c(x), z(x), v_i(x) \in \R^n$ and $\alpha(x), L(x), \beta_i(x) \in \R$ be defined as in Definition~\ref{def:basic_functions}. Let $K(x) \in \R^{n \times n}$ and $\wt{c}(x) \in \R^n$ be defined as in Definition~\ref{def:B1}.

Then, we define $B(x) \in \R^{n \times n}$ as follows:
    \begin{align*}
        B(x)
        := & ~\alpha(x)^{-2} \cdot \diag(z(x)) \cdot  K(x)^\top K(x) \cdot \diag(z(x)) \\
        - & ~ \alpha(x)^{-2} \cdot z(x) \cdot \wt{c}(x)^{\top} \cdot \diag (z(x)) - \alpha(x)^{-2} \\
        \cdot & ~ \diag (z(x)) \cdot \wt{c}(x) \cdot z(x)^{\top} \\
        + & ~ \alpha(x)^{-1} \cdot \mathrm{diag} ( \wt{c}(x) \circ u_2(x) ).
    \end{align*}
    
Furthermore, we defined $B_{\mat}(x), B_{\rank}(x), B_{\diag}(x) \in \R^{n \times n}$ as follows:
\begin{align*}
   B_{\mat}(x)
    := & ~ \alpha(x)^{-2} \cdot \diag(z(x)) \cdot  K(x)^\top K(x) \cdot \diag(z(x))\\ 
    B_{\rank}(x) 
    := & ~ \alpha(x)^{-2} \cdot( z(x) \cdot \wt{c}(x)^{\top} \cdot \diag (z(x)) \\
    + & ~ \diag (z(x)) \cdot \wt{c}(x) \cdot z(x)^{\top})  \\
    B_{\diag}(x) := & ~ \alpha(x)^{-1} \cdot  \diag( \wt{c}(x) \circ u_2(x) ),
\end{align*}
so that
\begin{align*}
    B(x) = B_{\mat}(x) - B_{\rank}(x) + B_{\diag}(x).
\end{align*}

\end{definition}

\section{Hessian is PSD}
\label{sec:hessian_PSD}

In this section, we mainly prove Lemma~\ref{lem:psd_lower_bound}.

\subsection{PSD Lower Bound}
\begin{lemma}\label{lem:psd_lower_bound}
Let $x \in \R^d$ be an arbitrary vector. Let $u_1(x), u_2(x), u(x), f(x), c(x), z(x), v_i(x) \in \R^n$ and $\alpha(x), L(x), \beta_i(x) \in \R$ be defined as in Definition~\ref{def:basic_functions}. Let $K(x), B(x), B_{\mat}(x), B_{\rank}(x), B_{\diag}(x) \in \R^{n \times n}$ and $\wt{c}(x) \in \R^n$ be defined as in Definition~\ref{def:B}. Let $\beta \in (0, 0.1)$ and $\beta < \alpha(x)$.

    Then, we have 
    \begin{itemize}
        \item Part 1. 
        \begin{align*}
            0 \preceq B_{\mat}(x) \preceq \beta^{-2} \cdot 16n^2 \exp(2R^2) \cdot I_n
        \end{align*}
        \item Part 2. 
        \begin{align*}
            -10\beta^{-2} n \exp(R^2)  \cdot I_n\preceq  - B_{\rank}(x) \preceq 10\beta^{-2} n \exp(R^2)  \cdot I_n
        \end{align*}
        \item Part 3. 
        \begin{align*}
            -4\beta^{-1} n \exp(R^2)   \cdot I_n  \preceq B_{\diag}(x) \preceq 4 \beta^{-1} n \exp(R^2)   \cdot I_n
        \end{align*}
        \item Part 4. 
        \begin{align*}
            -14 \beta^{-2}n\exp(R^2) \cdot I_n \preceq B(x) \preceq 30 \beta^{-2} n^2 \exp(2R^2)\cdot I_n
        \end{align*}
    \end{itemize}
\end{lemma}

\begin{proof}
    {\bf Proof of Part 1.}
    
    On the one hand,
    \begin{align*}
       B_{\mat}
        = & ~ \alpha(x)^{-2} \cdot \diag(z(x)) \cdot  K(x)^{\top} K(x) \cdot \diag(z(x)) \\
        \preceq & ~ \alpha(x)^{-2}\|\diag(z(x)) K(x)^{\top} \|^2 \cdot I_n \\
        \preceq & ~ \alpha(x)^{-2}\|\diag(z(x))\|^2 \| K(x)^{\top} \|^2 \cdot I_n \\
        \preceq & ~ \alpha(x)^{-2} \| z(x)\|_2^2 \cdot 4n \cdot I_n \\
        \preceq & ~ \beta^{-2} \cdot 16n^2 \exp(2R^2) \cdot I_n
    \end{align*}
where the first step follows from definition of $B_{\mat}$, the second step follows from {\bf Part 1} of Fact~\ref{fac:psd_rule}, the third step follows from {\bf Part 4} of Fact~\ref{fac:matrix_norm}, the fourth step follows from {\bf Part 2,4} of Fact~\ref{fac:vector_norm} and {\bf Part 7} of Lemma~\ref{lem:upper_bound}, and the final step follows from {\bf Part 8} of Lemma~\ref{lem:upper_bound} and $\alpha(x) > \beta$.

On the other hand, since $B_{\mat}$ is a positive semi-definite matrix, then $B_{\mat} \succeq 0$. 

{\bf Proof of Part 2}

On the one hand 
\begin{align*}
    B_{\rank}(x)  
    = & ~ \alpha(x)^{-2} \cdot z(x) \cdot \wt{c}(x)^{\top} \cdot \diag (z(x))\\
    + & ~ \alpha(x)^{-2} \cdot \diag (z(x)) \cdot \wt{c}(x) \cdot z(x)^{\top}\\
    \preceq & ~ \alpha(x)^{-2} \cdot (z(x) z(x)^{\top} \\
    + & ~ \wt{c}(x)^{\top}\cdot \diag(z(x)) \cdot (\wt{c}(x)^{\top}\cdot \diag(z(x)))^{\top}) \\ 
    \preceq & ~ \alpha(x)^{-2} (\|z(x) \|_2^2  + \|\wt{c}(x)^{\top}\diag(z(x))\|_2^2  ) \cdot I_n\\
    \preceq & ~ \alpha(x)^{-2} (2\sqrt{n} \exp(R^2) + \|\wt{c}(x) \|_2^2 \|z(x) \|_2^2) \cdot I_n\\
    \preceq & ~ \alpha(x)^{-2} (2\sqrt{n} \exp(R^2) + 8n \exp(R^2)) \cdot I_n \\
    \preceq & ~ 10 \beta^{-2}n \exp(R^2)  \cdot I_n
\end{align*}
where the first step follows from the definition of $B_{\rank}(x)$, the second step follows from {\bf Part 4} of Fact~\ref{fac:psd_rule}, the third step follows from {\bf Part 1} of Fact~\ref{fac:psd_rule}, the fourth step follows from {\bf Part 8} of Lemma~\ref{lem:upper_bound} and {\bf Part 9} of Fact~\ref{fac:vector_norm}, the fifth step follows from {\bf Part 8, 10} of Lemma~\ref{lem:upper_bound}, and the last step follows from $n > 1$ and $\alpha(x) > \beta$.

Then, by multiplying $-1$ on both sides, we can get
\begin{align*}
    - B_{\rank}(x) \succeq  -10 \beta^{-2}n \exp(R^2)  \cdot I_n
\end{align*}

On the other hand, the proof of the lower bound is similar to the previous one, so we omit it here.

{\bf Proof of Part 3}

On the one hand 
\begin{align*}
    B_{\diag}(x) 
    = & ~ \alpha(x)^{-1} \cdot  \diag( \wt{c}(x) \circ u_2(x) )\\
    \preceq & ~ \alpha(x)^{-1} \| \wt{c}(x) \|_2 \| u_2(x)\|_2 \cdot I_n\\
    \preceq & ~ 4 \beta^{-1} n \exp(R^2)   \cdot I_n 
\end{align*}
where the first step follows from the definition of $B_{\diag}(x)$, the second step follows from {\bf Part 7} of Fact~\ref{fac:psd_rule}, and the last step follows from {\bf Part 1, 10} of Lemma~\ref{lem:upper_bound} and $\alpha(x) > \beta$.

On the other hand, the proof of the lower bound is similar to the previous one, so we omit it here.

{\bf Proof of Part 4}

On the one hand 
\begin{align*}
    B(x)
    =  &  ~ B_{\mat}(x) - B_{\rank}(x)  + B_{\diag}(x) \\
    \preceq & ~  \beta^{-2} \cdot 16n^2 \exp(2R^2) \cdot I_n + 10 \beta^{-2}n \exp(R^2)  \cdot I_n \\
    + & ~ 4 \beta^{-1} n \exp(R^2)   \cdot I_n \\
    \preceq & ~ 30 \beta^{-2} n^2 \exp(2R^2)\cdot I_n 
\end{align*}
where the first step follows from Definition~\ref{def:B}, the second step follows {\bf Part 1, 2, 3}, and the last step follows from $\beta^{-1} > 1, n > 1$, and $\exp(2R^2) > \exp(R^2)$.

On the other hand, we have 
\begin{align*}
    B(x)
    =  &  ~ B_{\mat}(x) - B_{\rank}(x)  + B_{\diag}(x) \\
    \succeq & ~   - 10 \beta^{-2}n \exp(R^2)  \cdot I_n - 4 \beta^{-1} n \exp(R^2)   \cdot I_n) \\ 
    \succeq & ~ -14 \beta^{-2}n\exp(R^2) \cdot I_n
\end{align*}
where the first step follows from Definition~\ref{def:B}, the second step follows {\bf Part 1, 2, 3}, and the last step follows from $\beta^{-1} > 1$.
\end{proof}

\section{Hessian is Lipschitz}
\label{sec:hessian_lip}

In this section, we find the upper bound of $\| \nabla^2 L(x) - \nabla^2L(y)\|$ and thus proved that $\nabla^2L$ is
Lipschitz. More specifically, in section~\ref{sub:hessian_lip:main}, we give a summary of the main properties developed in this whole section. In Section~\ref{sub:hessian_lip:upper_bound}, we present the upper bound of the norms of the functions we analyzed before. In Section~\ref{sub:hessian_lip:lip}, we present the Lipschitz properties of the functions we analyzed before. In Section~\ref{sub:hessian_lip:summary}, we summarize the four steps of the Lipschitz for matrix functions. In Section~\ref{sub:hessian_lip:step1}, we analyze the first step of the Lipschitz for matrix function $\alpha(x)^{-2} \cdot \diag(z(x))^{\top}  \cdot  K(x)^\top K(x) \cdot \diag(z(x))$. In Section~\ref{sub:hessian_lip:step2}, we analyze the second step of the Lipschitz for matrix function $\alpha(x)^{-2} \cdot z(x) \cdot \wt{c}(x)^{\top} \cdot \diag (z(x))$. In Section~\ref{sub:hessian_lip:step3}, we analyze the third step of the Lipschitz for matrix function $\alpha(x)^{-2} \cdot \diag (z(x)) \cdot \wt{c}(x) \cdot z(x)^{\top}$. In Section~\ref{sub:hessian_lip:step4}, we analyze the fourth step of the Lipschitz for matrix function $\alpha(x)^{-1} \cdot \mathrm{diag} ( \wt{c}(x) \circ u_2(x) )$. 

\subsection{Main results}
\label{sub:hessian_lip:main}

In this section, we present the main lemma which is the summary of the properties developed in this whole section.

\begin{lemma}\label{lem:lips_main}

Let $H(x) = \frac{\d ^2 L}{\d x^2}$.

    Then we have 
    \begin{align*}
        \|H(x) - H(y) \|  \leq   20 \beta^{-5} n^{3.5} \exp(8R^2) \|x -y \|_2
    \end{align*}
\end{lemma}
\begin{proof}

The definition of $G_i$ is as follows
\begin{itemize}
    \item $G_1(x) = \alpha(x)^{-2} \cdot \diag(z(x))\cdot  K(x)^\top K(x) \cdot \diag(z(x))$
    \item $G_2(x) = -\alpha(x)^{-2} \cdot z(x) \cdot \wt{c}(x)^{\top} \cdot \diag (z(x))$
    \item $G_3(x) = -\alpha(x)^{-2} \cdot \diag (z(x)) \cdot \wt{c}(x) \cdot z(x)^{\top}$
    \item $G_4(x) = \alpha(x)^{-1} \cdot \mathrm{diag} ( \wt{c}(x) \circ u_2(x) )$,
\end{itemize}
which we define and analyze in Lemma~\ref{lem:lips_G_1}, \ref{lem:lips_G_2},  \ref{lem:lips_G_3}, \ref{lem:lips_G_4}, respectively.

Then, we have
    \begin{align*}
        \|H(x) - H(y) \| 
        = & ~ \|A \| \|\sum_{i=1}^4 G_i(x) - G_i(y) \|\|A\|\\
        \leq & ~ R^2 \cdot \|\sum_{i=1}^4 G_i(x) - G_i(y) \| \\
        \leq & ~ R^2 \cdot 20 \beta^{-5} n^{3.5} \exp(7R^2) \|x -y \|_2\\
        \leq & ~ 20 \beta^{-5} n^{3.5} \exp(8R^2) \|x -y \|_2
    \end{align*}
where the first step follows from the definition of $G_i$ (see Lemma~\ref{lem:summar_four_steps}) and matrix spectral norm, the second step follows from $\| A\|\leq R$, the second step follows from Lemma~\ref{lem:summar_four_steps}, and the last step follows from $R^2 \leq \exp(R^2)$
\end{proof}

\subsection{A core Tool: Upper Bound for Several 
Basic Functions}
\label{sub:hessian_lip:upper_bound}

In this section, we find the upper bound for the norms of the functions we analyze.

\begin{lemma}\label{lem:upper_bound}

Let $R \geq 4$. Let $A \in \R^{n \times d}$ and $x \in \R^d$ satisfy $\|A\| \leq R$ and $\|x\|_2 \leq R$. Let $b \in \R^{n}$ satisfy $\| b\|_1 \leq 1$. Let $u_1(x), u_2(x), u(x), f(x), c(x), z(x), v_i(x) \in \R^n$ and $\alpha(x), L(x), \beta_i(x) \in \R$ be defined as in Definition~\ref{def:basic_functions}. Let $K(x), B(x), B_{\mat}(x), B_{\rank}(x), B_{\diag}(x) \in \R^{n \times n}$ and $\wt{c}(x) \in \R^n$ be defined as in Definition~\ref{def:B}. Let $\beta \in (0, 0.1)$, and $\langle \exp(Ax) , {\bf 1}_n \rangle$, $\langle \exp(Ay) , {\bf 1}_n \rangle$, $\langle \exp(Ax)+ Ax, {\bf 1}_n \rangle$, and $\langle \exp(Ay)+ Ay, {\bf 1}_n \rangle$ be greater than or equal to $\beta$, respectively. Let $R_f = 2  \beta^{-1} \cdot (R\exp(R^2)+R)\cdot (n \cdot \exp(R^2) + \sqrt{n} \cdot R^2 )$. We define $R_{f, 2} \in \R$ as $R_{f, 2} : = 2 \sqrt{n} \beta^{-1} \exp(R^2)$.

    Then, we have 
    \begin{itemize}
        \item Part 1. $\|\exp(Ax) \|_2 \leq \sqrt{n}\exp(R^2) $
        \item Part 2. $\| \exp(Ax)+ Ax \|_2 \leq 2\sqrt{n} \exp(R^2)$
        \item Part 3. $ | \alpha(x)|   \geq \beta$
        \item Part 4. $ |\alpha(x)^{-1}| \leq \beta^{-1} $
        \item Part 5. $  \| f(x)\|_2 \leq R_{f, 2}$
        \item Part 6. $ \| c(x) \|_2  \leq 2 R_{f, 2}$
        \item Part 7. $ \| K(x) \|\leq 3 \sqrt{n} \cdot R_{f, 2}$
        \item Part 8 $ \| z(x) \|_2\leq 2 \sqrt{n} \cdot \exp(R^2) $
        \item Part 9 $ |\alpha(x)^{-2}| \leq \beta^{-2} $
        \item Part 10 $\|\wt{c}(x) \|_2 \leq 10 \sqrt{n} R_{f, 2}^2 $ 
      \end{itemize}
\end{lemma}

\begin{proof}
\ifdefined\isarxiv
{\bf Proof of Part 1}
    \begin{align*}
        \|\exp(Ax) \|_2 
        \leq & ~ \sqrt{n} \cdot  \|\exp(Ax) \|_\infty\\
        \leq & ~ \sqrt{n} \cdot  \exp(\|(Ax) \|_\infty)\\
        \leq & ~ \sqrt{n} \cdot  \exp(\|(Ax) \|_2)\\
        \leq & ~ \sqrt{n} \cdot  \exp(R^2)
    \end{align*}
    where the first step follows from Fact~\ref{fac:vector_norm}, the second step follows from {\bf Part 6} of Fact~\ref{fac:vector_norm},
the third step follows from {\bf Part 6} of Fact~\ref{fac:vector_norm}, and the last step follows from $\|A\| \leq R$ and $\| x\|_2 \leq R$.

    {\bf Proof of Part 2}
    \begin{align*}
        \|\exp(Ax) +Ax \|_2 
        \leq & ~ \|\exp(Ax) \|_2 + \|Ax \|_2 \\
        \leq & ~ \sqrt{n} \cdot  \exp(R^2) + R^2\\
        \leq & ~ 2\sqrt{n}\exp(R^2)
    \end{align*}
where the first step follows from {\bf Part 8} of Fact~\ref{fac:vector_norm}, the second step follows from {\bf Part 1} and $\|A\| \leq R, \| x\|_2 \leq R$, and the last step follows from $n > 1, \exp(R^2) \geq R^2$. 

    {\bf Proof of Part 3}
    \begin{align*}
        |\alpha(x)| 
        = & ~ | \langle u(x), {\bf 1}_n \rangle | \\
        \geq & ~ | \langle \exp(Ax) + Ax, {\bf 1}_n \rangle | \\
        \geq & ~ \beta
    \end{align*}
where the first step follows from the definition of $\alpha(x)$ (see Definition~\ref{def:basic_functions}), the second step follows the definition of $u(x)$ (see Definition~\ref{def:basic_functions}), and the last step follows from the assumption $\langle \exp(Ax)+ Ax, {\bf 1}_n\rangle \geq \beta$.

    {\bf Proof of Part 4}

   We have
   \begin{align*}
        |\alpha(x)^{-1}| 
        \leq & ~ |\beta^{-1}| \\
        \leq & ~  \beta^{-1} 
   \end{align*}
where the first step follows from {\bf Part 3} of Lemma~\ref{lem:upper_bound}, the second step follows from $\beta^{-1} > 0$.

    {\bf Proof of Part 5}

    First, we analyze the following equation:
    \begin{align}\label{eq:bound_for_expax_ax}
        |\langle \exp(Ax) + Ax, {\bf 1}_n \rangle^{-1}| 
        = & ~ |\langle \exp(Ax) + Ax, {\bf 1}_n \rangle|^{-1} \notag\\
        \leq & ~ \beta^{-1},
    \end{align}
    where the first step follows from simple algebra and the second step follows from the assumption in the Lemma statement.

    Then, we have
    \begin{align*}
        \|f(x)\|_2 
        = & ~ \|\alpha(x)^{-1} u(x)\|_2\\
        = & ~ \|\langle \exp(Ax) + Ax, {\bf 1}_n \rangle^{-1} (\exp(Ax) + Ax)\|_2\\
        = & ~ |\langle \exp(Ax) + Ax, {\bf 1}_n \rangle^{-1}| \cdot \|(\exp(Ax) + Ax)\|_2\\
        \leq & ~ \beta^{-1} \cdot \|(\exp(Ax) + Ax)\|_2\\
        \leq & ~ 2 \sqrt{n} \beta^{-1} \exp(R^2)\\
        = & ~ R_{f, 2},
    \end{align*}
    where the first step follows from the definition of $f(x)$ (see Definition~\ref{def:basic_functions}), the second step follows from the definition of $\alpha(x)$ and $u(x)$ (see Definition~\ref{def:basic_functions}), the third step follows from Fact~\ref{fac:vector_norm}, the fourth step follows from Eq.~\eqref{eq:bound_for_expax_ax}, and the fifth step follows from {\bf Part 2}, and the last step follows from the definition of $R_{f, 2}$.

    {\bf Proof of Part 6}
    \begin{align*}
        \|c(x)\|_2 
        = & ~ \| f(x) - b\|_2 \\
        \leq & ~ \| f(x)\|_2 + \| b\|_2 \\
        \leq & ~ R_{f, 2} + 1\\
        \leq & ~ 2R_{f, 2},
    \end{align*}
where the first step follows from the definition of $c(x)$ (see Definition~\ref{def:basic_functions}), the second step follows from {\bf Part 8} of Fact~\ref{fac:vector_norm}, the third step follows from {\bf Part 5} of Lemma~\ref{lem:upper_bound} and $\| b\|_2 \leq \| b\|_1 \leq 1$, and the last step follows from $R_{f,2} \geq 1$. 

    {\bf Proof of Part 7}
    \begin{align*}
        \|K(x) \|
        = & ~ \| (I_n - f(x) \cdot {\bf 1}_n^\top )\|  \\
        \leq & ~  \|I_n \| + \|f(x) \cdot {\bf 1}_n^\top \| \\
        \leq & ~ 1  + \|f(x) \|_2 \cdot \| {\bf 1}_n^\top\|_2 \\
        \leq & ~ 1 + 2 \sqrt{n} \beta^{-1} \exp(R^2) \cdot \sqrt{n}\\
        \leq & ~ 3 \sqrt{n} R_{f, 2},
    \end{align*}
where the first step follows from the definition of $K(x)$, the second step follows from the {\bf Part 3} of Fact~\ref{fac:matrix_norm}, the third step follows from $\|I_n \| =  1$ and {\bf Part 9} of Fact~\ref{fac:vector_norm}, and the fourth step follows from {\bf Part 5} of Lemma~\ref{lem:upper_bound}, and the last step follows from the simple algebra.

    {\bf Proof of Part 8}
    \begin{align*}
        \| z(x)\|_2
        = & ~ \| u_2(x) + {\bf 1}_n \| \\
        \leq & ~ \| u_2(x)\|_2 + \|{\bf 1}_n \|_2\\
        \leq & ~ \sqrt{n} \cdot (\exp(R^2) + 1) \\
        \leq & ~ 2 \sqrt{n} \exp(R^2)
    \end{align*}
where the first step follows from the the definition of $z(x)$ (see Definition~\ref{def:basic_functions}), the second step follows from {\bf Part 8} of Fact~\ref{fac:vector_norm}, the third step follows from {\bf Part 1} of Lemma~\ref{lem:upper_bound}, and the last step follows from Fact~\ref{fac:basic_algebraic_properties}.

{\bf Proof of Part 9}
    \begin{align*}
        | \alpha(x)^{-2}| 
        = & ~|\alpha(x)^{-1}  |^2 \\
        \leq & ~  \beta^{-2}
    \end{align*}
where the first step follows from simple algebra, and the last step follows from {\bf Part 4} of Lemma~\ref{lem:upper_bound}.

{\bf Proof of Part 10}
\begin{align*}
    \| \wt{c}(x) \|_2 
    = & ~\|K(x)^{\top} c(x) \|_2 \\
    \leq & ~ \|K(x) \| \| c(x) \|_2 \\
    \leq & ~ 3 \sqrt{n} R_{f, 2} \cdot 2R_{f, 2}\\
    \leq & ~ 10 \sqrt{n} R_{f, 2}^2 ,
\end{align*}
where the first step follows from Definition of $\wt{c}(x)$, the second step follows from {\bf Part 7} of Fact~\ref{fac:matrix_norm}, the third step follows from {\bf Part 6 and 7} of Lemma~\ref{lem:upper_bound}, and the last step follows from simple algebra.
\else
The proof details can be found in full version \cite{full}.
\fi
\end{proof}

\subsection{A core Tool: Lipschitz Property for Several 
Basic Functions}
\label{sub:hessian_lip:lip}

In this section, we present the Lipschitz property for the functions we analyze.

\begin{lemma}[Basic Functions Lipschitz Property]\label{lem:basic_lips}

Let $R \geq 4$. Let $A \in \R^{n \times d}$ and $x \in \R^d$ satisfy $\|A\| \leq R$ and $\|x\|_2 \leq R$. Let $b \in \R^{n}$ satisfy $\| b\|_1 \leq 1$. Let $u_1(x), u_2(x), u(x), f(x), c(x), z(x), v_i(x) \in \R^n$ and $\alpha(x), L(x), \beta_i(x) \in \R$ be defined as in Definition~\ref{def:basic_functions}. Let $K(x), B(x), B_{\mat}(x), B_{\rank}(x), B_{\diag}(x) \in \R^{n \times n}$ and $\wt{c}(x) \in \R^n$ be defined as in Definition~\ref{def:B}. Let $\beta \in (0, 0.1)$, and $\langle \exp(Ax) , {\bf 1}_n \rangle$, $\langle \exp(Ay) , {\bf 1}_n \rangle$, $\langle \exp(Ax)+ Ax, {\bf 1}_n \rangle$, and $\langle \exp(Ay)+ Ay, {\bf 1}_n \rangle$ be greater than or equal to $\beta$, respectively. Let $R_f =  6  \beta^{-2} \cdot n \cdot  \exp(3R^2)$.

    Then, we have 
    \begin{itemize}
        \item Part 1. $\|Ax - Ay\|_2 \leq R \cdot \|x -y \|_2$
        \item Part 2. $\|\exp(Ax) - \exp(Ay) \|_2 \leq R \exp(R^2) \cdot \|x -y  \|_2 $
        \item Part 3. $|\alpha(x) - \alpha(y) | \leq 2 \sqrt{n}R\exp(R^2)\|x -y \|_2$
        \item Part 4. $| \alpha(x)^{-1} - \alpha(y)^{-1}| \leq \beta^{-2} \cdot |\alpha(x) - \alpha(y)| $
        \item Part 5. $\|f(x) -f(y) \|_2 \leq R_f
        \cdot \|x -y  \|_2$
        \item Part 6. $\|c(x) -c(y) \|_2 \leq R_f
        \cdot \|x -y  \|_2$
        \item Part 7. $\|z(x) -z(y) \|_2 \leq R\exp(R^2)\|x -y \|_2$
        \item  Part 8. $\|K(x) - K(y) \| \leq   \sqrt{n}\cdot R_f \cdot \|x -y \|_2 $ 
        \item Part 9. $\|\diag(z(x)) - \diag(z(y)) \| \leq  R \exp(R^2)\|x -y \|_2$
        \item Part 10.
        $| \alpha(x)^{-2} - \alpha(y)^{-2} |  \leq  4\beta^{-3} \sqrt{n}R \exp(R^2) \cdot \|x -y \|_2 $
        \item  Part 11. $\| \wt{c}(x) - \wt{c}(y) \|_2 \leq 5\sqrt{n}\cdot R_f \cdot R_{f,2}\cdot \|x -y \|_2$
        \item  Part 12. $ \|\diag(\wt{c}(x) \circ u_2(x)) - \diag(\wt{c}(y) \circ u_2(y))  \| \leq 10 n \cdot R_f \cdot R_{f,2} \cdot \exp(2R^2)  \|x -y \|_2$
    \end{itemize}
\end{lemma}

\begin{proof}
\ifdefined\isarxiv

    {\bf Proof of Part 1}
    \begin{align*}
        \|Ax - Ay\|_2
        \leq & ~ \| A\| \| x -y\|_2\\
        \leq & ~  R \cdot \|x -y \|_2
    \end{align*}
where the first step follows from {\bf Part 7} of Fact~\ref{fac:matrix_norm}, and the last step follows from $\|A\| \leq R$.

    {\bf Proof of Part 2}
    \begin{align*}
        \|\exp(Ax) - \exp(Ay) \|_2 
        \leq & ~ \exp(R^2) \|Ax -Ay \|_2 \\
        \leq & ~ \exp(R^2) \| A\|\|x -y \|_2 \\
        \leq & ~ R\exp(R^2) \|x -y \|_2  
    \end{align*}
    where the first step follows from {\bf Part 10} of Fact~\ref{fac:vector_norm}, the second step follows from {\bf Part 7} of Fact~\ref{fac:matrix_norm}, the third step follows from $\| A \| \leq R$.

    {\bf Proof of Part 3}
    \begin{align*}
        & ~ |\alpha(x) -\alpha(y) | \\
        = & ~ |\langle (\exp(Ax) + Ax)-  (\exp(Ay)+Ay), {\bf 1}_n\rangle | \\
        \leq & ~ \| (\exp(Ax) + Ax)-  (\exp(Ay)+Ay)\|_2 \cdot \sqrt{n}\\
        \leq & ~(\|\exp(Ax)- \exp(Ay)\|_2 +\|Ax- Ay\|_2 ) \cdot \sqrt{n} \\
        \leq & ~\sqrt{n}( R\exp(R^2) + R) \cdot \|x - y \|_2 \\
        \leq & ~ 2 \sqrt{n}R \exp(R^2) \cdot \|x -y \|_2
    \end{align*} 
where the first step follows from the definition of $\alpha(x)$ (see Definition~\ref{def:basic_functions}), the second step follows from {\bf Part 1} of Fact~\ref{fac:vector_norm} (Cauchy-Schwarz inequality),  the third step follows from {\bf Part 8} of Fact~\ref{fac:vector_norm}, the fourth step follows from {\bf Part 1 and 2} of Lemma~\ref{lem:basic_lips}, and the last step follows from {\bf Part 1} of  Fact~\ref{fac:basic_algebraic_properties}.

    {\bf Proof of Part 4}
    \begin{align*}
        |\alpha(x)^{-1} -\alpha(y)^{-1} | 
        = & ~ \alpha(x)^{-1}\cdot \alpha(y)^{-1} |\alpha(x) -\alpha(y) | \\
        \leq & ~\beta^{-2}
        \cdot |\alpha(x) -\alpha(y)|
    \end{align*}
where the first step follows from the simple algebra, and the last step follows from $\alpha(x), \alpha(y) \geq \beta$.

    {\bf Proof of Part 5}
    \begin{align}\label{eq:fx-fy}
        & ~ \| f(x) - f(y)\|_2 \notag\\
        = & ~\|\alpha(x)^{-1} \cdot (\exp(x)+ Ax) - \alpha(y)^{-1} \cdot (\exp(y)+ Ay)  \|_2 \notag\\
        \leq & ~ \|\alpha(x)^{-1} \cdot (\exp(x)+ Ax) - \alpha(x)^{-1} \cdot (\exp(y)+ Ay)  \|_2  \notag\\
        + & ~ \|\alpha(x)^{-1} \cdot (\exp(y)+ Ay) - \alpha(y)^{-1} \cdot (\exp(y)+ Ay)  \|_2 \notag\\
        \leq & ~ \alpha(x)^{-1} \cdot \| (\exp(x)+ Ax) -(\exp(y)+ Ay) \|_2 \notag\\
        + & ~ |\alpha(x)^{-1}  -\alpha(y)^{-1}  |\| \exp(Ay )+ Ay \|,
    \end{align}
where the first step follows from the definition of $f(x)$ and $u(x)$ (see Definition~\ref{def:basic_functions}), the second step follows from triangle inequality ({\bf Part 8} of Fact~\ref{fac:vector_norm}), and the last step follows from {\bf Part 7} of Fact~\ref{fac:vector_norm}.

    For the first term in the above, we have
    \begin{align}\label{eq:fx-fypart1}
        & ~ \alpha(x)^{-1} \cdot \| (\exp(x)+ Ax) -(\exp(y)+ Ay) \|_2 \notag \\
        \leq & ~ \beta^{-1} \cdot \| (\exp(x)+ Ax) -(\exp(y)+ Ay) \|_2  \notag\\
        \leq & ~\beta^{-1} \cdot (\| \exp(x) -\exp(y) \|_2 + \| Ax -Ay \|_2) \notag\\
        \leq & ~ \beta^{-1} \cdot (R\exp(R^2) \|x -y \|_2 + R \cdot \|x -y \|_2) \notag \\
        = & ~ \beta^{-1} \cdot (R\exp(R^2)+R)\cdot \|x -y \|_2 \notag \\
        \leq & ~ 2\beta^{-1} \cdot  R \exp(R^2) \cdot \| x- y \|_2
    \end{align}
where the first step follows from $\alpha(x) \geq \beta $, the second step follows from {\bf Part 8} of Fact~\ref{fac:vector_norm}, the third step follows from {\bf Part 1 and Part 2} of Lemma~\ref{lem:basic_lips}, the fourth step follows from simple algebra, and the last step follows from Fact~\ref{fac:basic_algebraic_properties}.  

For the second term in the above, we have
 \begin{align}\label{eq:fx_fypart2}
     & ~ |\alpha(x)^{-1}  -\alpha(y)^{-1}  |\| \exp(Ay )+ Ay \|_2 \notag \\
     \leq &  ~\beta^{-2}
    \cdot |\alpha(x) -\alpha(y) | \cdot \| \exp(Ay )+ Ay \|_2 \notag \\
    \leq & ~ \beta^{-2}
        \cdot |\alpha(x) -\alpha(y) | \cdot  2 \sqrt{n} \exp(R^2)\notag \\
    \leq & ~ \beta^{-2}
        \cdot 2R \exp(R^2)\cdot \|x -y \|_2 \cdot \sqrt{n}  \cdot 2\sqrt{n} \exp(R^2)  \notag\\
    = & ~ 4\beta^{-2}
      \cdot R \cdot n\exp(2R^2) \cdot \|x -y \|_2 
 \end{align}
where the first step follows from the result of {\bf Part 4} of Lemma~\ref{lem:basic_lips}, the second step follows from the result of {\bf Part 2} of Lemma~\ref{lem:upper_bound}, the third step follows from the result of {\bf Part 3} of Lemma~\ref{lem:basic_lips}, and the last step follows from simple algebra.

Combining Eq.~\eqref{eq:fx-fy}, Eq.~\eqref{eq:fx-fypart1}, and Eq.~\eqref{eq:fx_fypart2} together, we have 
\begin{align*}
     \| f(x) - f(y)\|_2
     \leq & ~ 2 \beta^{-1} \cdot R\exp(R^2) \cdot \|x -y \|_2\\
    + & ~ 4  \beta^{-2}
        \cdot n \cdot R \exp(2R^2)\cdot \|x -y \|_2 \\
       \leq & ~ 6  \beta^{-2}
        \cdot n \cdot  \exp(3R^2)\cdot  \|x -y \|_2
\end{align*}
where the last step follows from $\beta^{-1} \geq 1$, $n \geq 1$, $R \geq 4$, and Fact~\ref{fac:basic_algebraic_properties}.  

    {\bf Proof of Part 6}
    \begin{align*}
        \|c(x) - c(y) \|_2 
        = & ~ \| f(x) - f(y) \|_2 \\
        \leq & ~ R_f \cdot \|x -y \|_2
    \end{align*}
where the first step follows from the definition of $c(x)$ (see Definition~\ref{def:basic_functions}), and the last step follows from {\bf Part 5} of Lemma~\ref{lem:basic_lips}.

    {\bf Proof of Part 7}
    \begin{align*}
        \|z(x)  - z(y)\|_2
        = & ~\| u_2(x) + {\bf 1}_n -u_2(y) - {\bf 1}_n  \| \\
        = & ~\| u_2(x) - u_2(y)\|_2 \\
        \leq & ~ R\exp(R^2)\|x -y \|_2
    \end{align*}
    where the first step follows from the definition of $z(x)$ (see Definition~\ref{def:basic_functions}), the second step follows from simple algebra, and the last step follows from the definition of $u_2(x)$ (see Definition~\ref{def:basic_functions}) and {\bf Part 2} of Lemma~\ref{lem:basic_lips}.

    {\bf Proof of Part 8}
    \begin{align*}
    \|K(x) -K(y) \| 
    = & ~ \| (I_n - f(x) \cdot {\bf 1}_n^\top ) -(I_n - f(y) \cdot {\bf 1}_n^\top )\| \\
    = & ~ \|  - (f(x) -f(y)) \cdot {\bf 1}_n^\top  \| \\
    \leq & ~ \|f(x) -f(y) \|_2 \cdot \|{\bf 1}_n^\top \|_2 \\
    \leq & ~  \sqrt{n}\cdot R_f \cdot \|x -y \|_2
    \end{align*}
where the first step follows from the definition of $K(x)$, the second step follows from the simple algebra, the third step follows from {\bf Part 9} of Fact~\ref{fac:vector_norm}, and the last step follows from {\bf Part 5} of Lemma~\ref{lem:basic_lips}.

{\bf Proof of Part 9}
\begin{align*}
    \| \diag(z(x)) - \diag(z(y)) \|
    = & ~ \| \diag(z(x) - z(y)) \| \\
    \leq & ~ \|z(x) - z(y) \|_{\infty} \\
    \leq & ~ \|z(x) - z(y) \|_2\\
    \leq & ~  R\exp(R^2)\|x -y \|_2
\end{align*}
where the first step follows from the simple algebra, the second step follows from {\bf Part 2} of Fact~\ref{fac:vector_norm}, the third step follows from {\bf Part 4} of Fact~\ref{fac:vector_norm}, and the last step follows from {\bf Part 7} of Lemma~\ref{lem:basic_lips}.

{\bf Proof of Part 10}
\begin{align*}
    | \alpha(x)^{-2} -  \alpha(y)^{-2} |
    = & ~  | (\alpha(x)^{-1} - \alpha(y)^{-1})(\alpha(x)^{-1} + \alpha(y)^{-1})| \\
    \leq & ~ |\alpha(x)^{-1} - \alpha(y)^{-1}|| \alpha(x)^{-1} + \alpha(y)^{-1}|\\
    \leq & ~ \beta^{-2} \cdot|\alpha(x) - \alpha(y) | \cdot |2\beta^{-1}|\\
    \leq & ~ 2\beta^{-3}|\alpha(x) - \alpha(y) |\\
    \leq & ~ 
    4\beta^{-3} \sqrt{n}R \exp(R^2) \cdot \|x -y \|_2
\end{align*}
where the first step follows from the simple algebra, the second step follows from the simple algebra, the third step follows from {\bf Part 4} of Lemma~\ref{lem:upper_bound} and {\bf Part 4} of Lemma~\ref{lem:basic_lips}, the fourth step follows from the simple algebra, and the last step follows from {\bf Part 3} of Lemma~\ref{lem:basic_lips}.

{\bf Proof of Part 11}
\begin{align*}
    & ~ \|\wt{c}(x) - \wt{c}(y) \|_2 \\
    =  & ~ \| K(x)^{\top} c(x) - K(y)^{\top} c(y) \| \\
    \leq & ~ \| K(x)^{\top} c(x) - K(y)^{\top} c(x) \| + \| K(y)^{\top} c(x) - K(y)^{\top} c(y) \| \\
    \leq & ~ \|K(x)^{\top} - K(y)^{\top}\|\cdot \| c(x)\|_2 + \|K(y)^{\top} \|\cdot \|c(x) - c(y) \|_2 \\
    \leq & ~\sqrt{n} R_f \cdot \|x -y \|_2 \cdot 2 R_{f,2} + 3 \sqrt{n} \cdot R_{f,2}\cdot R_f \cdot \|x -y \|_2 \\
    \leq & ~5\sqrt{n}\cdot R_f \cdot R_{f,2}\cdot \|x -y \|_2
\end{align*}
where the first step follows from the definition of $\wt{c}(x)$, the second step follows from the triangle inequality, the third step follows from {\bf Part 7} of Fact~\ref{fac:matrix_norm}, the fourth step follows from {\bf Part 6, 8} of Lemma~\ref{lem:basic_lips} and {\bf Part 6, 7} of Lemma~\ref{lem:upper_bound}, and the last step follows from the simple algebra.

{\bf Proof of Part 12}
\begin{align*}
    & ~ \|\diag(\wt{c}(x) \circ u_2(x)) - \diag(\wt{c}(y) \circ u_2(y))  \|\\
    \leq &~\| \diag(\wt{c}(x)) \diag(u_2(x)) - \diag(\wt{c}(y)) \diag(u_2(y))\| \\
    \leq & ~ \|\diag(\wt{c}(x)) \diag(u_2(x)) - \diag(\wt{c}(x)) \diag(u_2(y)) \| \\
    + & ~\|\diag(\wt{c}(y)) \diag(u_2(x)) - \diag(\wt{c}(y)) \diag(u_2(y)) \| \\
    \leq & ~ \| \diag(\wt{c}(x)) - \diag(\wt{c}(y))\|\|\diag(u_2(x))\|\\
    + & ~ \|\diag(\wt{c}(y)) \|\| \diag(u_2(x)) - \diag(u_2(y))\| \\
    \leq & ~  \|\wt{c}(x) - \wt{c}(y)\|_2 \cdot \|u_2(x)\|_2 +\| \wt{c}(y)\|_2 \cdot \|u_2(x) - u_2(y)\|_2 \\
    \leq & ~ 5\sqrt{n}  \cdot R_{f,2} \cdot R_f \|x -y \|_2 \sqrt{n} \exp(R^2)  + 4\sqrt{n} R \exp(R^2)\|x -y \|_2 \\
    \leq & ~10 n \cdot R_f \cdot R_{f,2} \cdot \exp(2R^2)  \|x -y \|_2
    \end{align*}
where the first step follows from Fact~\ref{fac:circ_rules}, the second step follows from triangle inequality, the third step follows from Fact~\ref{fac:matrix_norm}, the fourth step follows from Fact~\ref{fac:vector_norm}, the fifth step follows from {\bf Part 2,11} of Lemma~\ref{lem:basic_lips} and {\bf Part 1,10} of Lemma~\ref{lem:upper_bound},  and the last step follows from $R_f > 1$, $R_{f,2} >1$, $n > 1$, $\beta^{-1} > 1$, and Fact~\ref{fac:basic_algebraic_properties}.

\else
The proof details can be found in full version \cite{full}.
\fi
\end{proof}

\subsection{Summary of Four Steps}
\label{sub:hessian_lip:summary}

In this section, we summarize the four steps which are discussed in the next four sections, respectively.

\begin{lemma}\label{lem:summar_four_steps}
    If the following conditions hold
    \begin{itemize}
        \item $G_1(x) = \alpha(x)^{-2} \cdot \diag(z(x))\cdot  K(x)^\top K(x) \cdot \diag(z(x))$
        \item $G_2(x) = -\alpha(x)^{-2} \cdot z(x) \cdot \wt{c}(x)^{\top} \cdot \diag (z(x))$
        \item $G_3(x) = -\alpha(x)^{-2} \cdot \diag (z(x)) \cdot \wt{c}(x) \cdot z(x)^{\top}$
        \item $G_4(x) = \alpha(x)^{-1} \cdot \mathrm{diag} ( \wt{c}(x) \circ u_2(x) )$
    \end{itemize}

    Then, we have 
    \begin{align*}
        \sum_{i = 1}^4 \|G_i(x) - G_i(y) \| \leq 20 \beta^{-5} n^{3.5} \exp(7R^2) \|x -y \|_2
    \end{align*}
\end{lemma}
\begin{proof}
    \begin{align*}
        \sum_{i = 1}^4 \|G_i(x) - G_i(y) \|
        \leq & ~ 5 \beta^{-5} n^{3.5}  \exp(7R^2)\cdot \|x -y \|_2\\
        + & ~   4 \beta^{-5}n^3\exp(7R^2) \cdot \|x -y \|_2 \\
        + & ~   4 \beta^{-5}n^3\exp(7R^2) \cdot \|x -y \|_2 \\
        + & ~   2 \beta^{-4}  n^3 \cdot   \exp(7R^2) \|x -y \|_2  \\
        \leq & ~ 20 \beta^{-5} n^{3.5} \exp(7R^2) \|x -y \|_2
    \end{align*}
where the first step follows from Lemma~\ref{lem:lips_G_1}, \ref{lem:lips_G_2},  \ref{lem:lips_G_3}, \ref{lem:lips_G_4}, the last step follows from $\beta^{-1} > 1, n > 1,R > 4$. 
\end{proof}

\subsection{  Calculation: Step 1 Lipschitz for Matrix Function \texorpdfstring{$\alpha(x)^{-2} \cdot \diag(z(x))^{\top}  \cdot  K(x)^\top K(x) \cdot \diag(z(x))$}{}}
\label{sub:hessian_lip:step1}

In this section, we analyze the first step, namely the Lipschitz for the matrix function $\alpha(x)^{-2} \cdot \diag(z(x))^{\top}  \cdot  K(x)^\top K(x) \cdot \diag(z(x))$.

\begin{lemma}\label{lem:lips_G_1}
    Let $G_1(x) = \alpha(x)^{-2} \cdot \diag(z(x))  \cdot  K(x)^\top K(x) \cdot \diag(z(x)) $.
    
    Then we have
    \begin{align*}
       \| G_1(x) - G_1(y) \| \leq 5 \beta^{-5} n^{3.5}  \exp(7R^2)\cdot \|x -y \|_2
    \end{align*}
\end{lemma}
\begin{proof}

\ifdefined\isarxiv
We define
\begin{align*}
    G_{1,1} := & ~ \alpha(x)^{-2}  \diag(z(x))    K(x)^\top K(x)  \diag(z(x)) \\
    - & ~ \alpha(y)^{-2}  \diag(z(x))   K(x)^\top K(x)  \diag(z(x)) \\
    G_{1,2} := & ~ \alpha(y)^{-2}  \diag(z(x))    K(x)^\top K(x)  \diag(z(x)) \\
    - & ~ \alpha(y)^{-2}  \diag(z(y))    K(x)^\top K(x)  \diag(z(x))\\
    G_{1,3} := & ~ \alpha(y)^{-2}  \diag(z(y))   K(x)^\top K(x)  \diag(z(x)) \\
    - & ~ \alpha(y)^{-2}  \diag(z(y))    K(y)^\top K(x)  \diag(z(x))\\
    G_{1,4} := & ~ \alpha(y)^{-2}  \diag(z(y))   K(y)^\top K(x)  \diag(z(x)) \\
    - & ~ \alpha(y)^{-2}  \diag(z(y))    K(y)^\top K(y)  \diag(z(x))\\
    G_{1,5} := & ~ \alpha(y)^{-2}  \diag(z(y))    K(y)^\top K(y)  \diag(z(x)) \\
    - & ~ \alpha(y)^{-2}  \diag(z(y))    K(y)^\top K(y)  \diag(z(y))
\end{align*}

we have 
\begin{align*}
    G_{1} =  G_{1,1} +  G_{1,2} + G_{1,3} +  G_{1,4}+  G_{1,5}
\end{align*}

Let's prove the $G_{1,1}$ first
\begin{align*}
    & ~ \|G_{1,1}\| \\
    = & ~ \| \alpha(x)^{-2}  \diag(z(x))   K(x)^\top K(x)  \diag(z(x)) \\
    - & ~ \alpha(y)^{-2}  \diag(z(x))  K(x)^\top K(x)  \diag(z(x)) \| \\
    \leq & ~ |\alpha(x)^{-2} - \alpha(y)^{-2}| 
    \cdot \|\diag(z(x))   K(x)^\top K(x)  \diag(z(x)  \| \\
    \leq & ~ |\alpha(x)^{-2} - \alpha(y)^{-2}| \\
    \cdot & ~ \| \diag( z(x) ) \| \cdot \| K(x)^\top  \| \| K(x) \| \cdot \| \diag( z(x) ) \| \\
    \leq & ~  |\alpha(x)^{-2} - \alpha(y)^{-2}| \cdot \| z(x) \|_{\infty}^2 \cdot \| K(x) \|^2  \\
    \leq & ~ 4\beta^{-3} \sqrt{n}R \exp(R^2) \cdot \|x -y \|_2 \cdot \|z(x) \|_2^2 \cdot (3 \sqrt{n} R_{f, 2})^2\\
    \leq & ~ 4\beta^{-3} \sqrt{n}R \exp(R^2) \cdot \|x -y \|_2 \cdot (2 \sqrt{n} \exp(R^2))^2 \cdot (3 \sqrt{n} R_{f, 2})^2 \\
    \leq & ~  200 \beta^{-3}  n^{2.5} \exp(4R^2) \cdot( 2 \sqrt{n} \beta^{-1} \exp(R^2))^2 \|x -y \|_2 \\
    \leq & ~ \beta^{-5} n^{3.5} \exp(7R^2) \|x-y \|_2
\end{align*}
where the first step follows from the definition of $G_{1,1}$, the second step follows from {\bf Part 6} of Fact~\ref{fac:matrix_norm}, the third step follows from {\bf Part 4} of Fact~\ref{fac:matrix_norm}, the fourth step follows from {\bf Part 2} of Fact~\ref{fac:vector_norm}, the fifth step follows from {\bf Part 4} of Fact~\ref{fac:vector_norm}, {\bf Part 10} of Lemma~\ref{lem:basic_lips}, and {\bf Part 7} of Lemma~\ref{lem:upper_bound}, the sixth step follows from {\bf Part 8} of Lemma~\ref{lem:upper_bound}, the seventh step follows from the definition of $R_{f,2}$, and the last step follows from simple algebra.

Then let's prove the $G_{1,2}$
\begin{align*}
     & ~ \|G_{1,2}\| \\
     = & ~ \| \alpha(y)^{-2}  \diag(z(x))    K(x)^\top K(x)  \diag(z(x)) \\
     - & ~ \alpha(y)^{-2}  \diag(z(y))    K(x)^\top K(x)  \diag(z(x)) \|\\
     \leq & ~ \|\diag(z(x)) - \diag(z(y))\| \\
     \cdot & ~ \| \alpha(y)^{-2}  K(x)^\top K(x)  \diag(z(x))\| \\
     \leq & ~ R \exp(R^2)\|x-y \|_2 \cdot |\alpha(y)^{-2} | \| K(x)^\top  \| \| K(x) \| \cdot \| \diag( z(x) ) \| \\
     \leq & ~ R \exp(R^2)\|x-y \|_2  \cdot \beta^{-2} \cdot 2 \sqrt{n} \exp(R^2) \cdot (3 \sqrt{n} R_{f, 2})^2\\
     \leq & ~ 24\beta^{-2} \cdot n^{1.5} \cdot  R \exp(3R^2) \cdot ( 2 \sqrt{n} \beta^{-1} \exp(R^2))^2\|x-y \|_2  \\
     \leq & ~  \beta^{-4} n^{2.5} \exp(6R^2) \|x- y \|_2
\end{align*}
where the first step follows from the Definition of $G_{1,2}$, the second step follows from the {\bf Part 4} of Fact~\ref{fac:matrix_norm}, the third step follows from {\bf Part 4 and 6} of Fact~\ref{fac:matrix_norm} and {\bf Part 9} of Lemma~\ref{lem:basic_lips}, the fourth step follows from the {\bf Part 7, 8, 9} of Lemma~\ref{lem:upper_bound}, the fifth step follows from the definition of $R_{f,2}$, and the last step follows from simple algebra.

Let's prove the  $G_{1,3}$
\begin{align*}
      & ~ \|G_{1,3}\| \\
      = & ~\|  \alpha(y)^{-2}  \diag(z(y))   K(x)^\top K(x)  \diag(z(x)) \\
      - & ~ \alpha(y)^{-2}  \diag(z(y))    K(y)^\top K(x)  \diag(z(x)) \| \\
      \leq & ~ \|K(x)^{\top} - K(y)^{\top} \| \cdot \| \alpha(y)^{-2}  \diag(z(y))  K(x)  \diag(z(x)) \| \\
      \leq & ~  \|K(x)^{\top} - K(y)^{\top} \| |\alpha(y)^{-2}|\cdot \| \diag( z(y) ) \| \cdot  \| K(x) \| \\
      \cdot & ~ \| \diag( z(x) ) \| \\
      \leq & ~ \sqrt{n}\cdot R_f \cdot \|x -y \|_2 \cdot  \beta^{-2} \cdot \|z(y) \|_2 \cdot \|z(x) \|_2 \cdot 3\sqrt{n} R_{f,2}\\
      \leq & ~ 3 n \cdot \beta^{-2} \cdot R_f \cdot R_{f,2} \cdot (2\sqrt{n}\cdot \exp(R^2))^2 \cdot \|x -y \|_2 \\
      \leq & ~ 12 \beta^{-2} \cdot  n^2  \cdot \exp(2R^2) \cdot (6  \beta^{-2}
        \cdot n \cdot  \exp(3R^2)) \cdot (2 \sqrt{n} \beta^{-1} \exp(R^2))\cdot \|x -y \|_2 \\
      \leq & ~ \beta^{-5} \cdot  n^{3.5}    \cdot  \exp(7R^2)\cdot \|x -y \|_2
\end{align*}
where the first step follows from the Definition of $G_{1,3}$, the second step follows from {\bf Part 4} of Fact~\ref{fac:matrix_norm}, the third step follows from {\bf Part 4, 6} of Fact~\ref{fac:matrix_norm}, the fourth step follows from {\bf Part 8} of Lemma~\ref{lem:basic_lips}, {\bf Part 9} of Lemma~\ref{lem:upper_bound}, and {\bf Part 3} of Fact~\ref{fac:vector_norm},  the fifth step follows from {\bf Part 8} of Lemma~\ref{lem:upper_bound}, the sixth step follows from the definition of $R_f, R_{f,2}$, and the last step follows from the simple algebra.

Proof of $G_{1,4}$ is similar to $G_{1,3}$, and the proof of $G_{1,5}$ is similar to $G_{1,2}$, so we omit them.

Then, by combining all results we get
\begin{align*}
    \|G_{1}(x) - G_1(y) \| 
    = & ~\|G_{1,1} + G_{1,2} + G_{1,3} + G_{1,4} + G_{1,5}\| \\
    \leq & ~ \beta^{-5} n^{3.5} \exp(7R^2) \|x-y \|_2 \\
    + & ~2 \beta^{-4} n^{2.5} \exp(6R^2) \|x- y \|_2\\
    + & ~ 2 \beta^{-5}  n^{3.5}     \exp(7R^2) \|x -y \|_2 \\
    \leq & ~ 5 \beta^{-5} n^{3.5}  \exp(7R^2)\cdot \|x -y \|_2
\end{align*}
where the first step follows from the Definitions of $G_{1,1}, G_{1,2}, G_{1,3}, G_{1,4}, G_{1,5}$, the second step follows from previous results, and the last step follows from simple algebra

\else
The proof details can be found in full version \cite{full}.
\fi
\end{proof}

\subsection{  Calculation: Step 2 Lipschitz for Matrix Function \texorpdfstring{$\alpha(x)^{-2} \cdot z(x) \cdot \wt{c}(x)^{\top} \cdot \diag (z(x))$}{}}
\label{sub:hessian_lip:step2}

In this section, we analyze the second step, namely the Lipschitz for the matrix function $\alpha(x)^{-2} \cdot z(x) \cdot \wt{c}(x)^{\top} \cdot \diag (z(x))$.

\begin{lemma}\label{lem:lips_G_2}
Let $G_2(x) = \alpha(x)^{-2} \cdot z(x) \cdot \wt{c}(x)^{\top} \cdot \diag (z(x))$. 

    Then we have
    \begin{align*}
        \| G_2(x) - G_2(y)\| \leq  4 \beta^{-5}n^3\exp(7R^2) \cdot \|x -y \|_2
    \end{align*}
\end{lemma}
\begin{proof}
    We define 
    \begin{align*}
        G_{2,1}
        := & ~  -(\alpha(x)^{-2} \cdot z(x) \cdot \wt{c}(x)^{\top} \cdot \diag (z(x)) \\
        - & ~ \alpha(y)^{-2} \cdot z(x) \cdot \wt{c}(x)^{\top} \cdot \diag (z(x)))\\
        G_{2,2}:= & ~-(  \alpha(y)^{-2} \cdot z(x) \cdot \wt{c}(x)^{\top} \cdot \diag (z(x)) \\
        - & ~ \alpha(y)^{-2} \cdot z(y) \cdot \wt{c}(x)^{\top} \cdot \diag (z(x)) )\\
        G_{2,3}:= & ~  -(\alpha(y)^{-2} \cdot z(y) \cdot \wt{c}(x)^{\top} \cdot \diag (z(x)) \\
        - & ~ \alpha(y)^{-2} \cdot z(y) \cdot \wt{c}(y)^{\top} \cdot \diag (z(x)))\\
        G_{2,4}:= & ~  - (\alpha(y)^{-2} \cdot z(y) \cdot \wt{c}(y)^{\top} \cdot \diag (z(x)) \\
        - & ~ \alpha(y)^{-2} \cdot z(y) \cdot \wt{c}(y)^{\top} \cdot \diag (z(y)))\\
    \end{align*}

    Then let's prove $G_{2,1}$ first
    \begin{align*}
        & ~ \|G_{2,1}\| \\
        = & ~ \| -(\alpha(x)^{-2} \cdot z(x) \cdot \wt{c}(x)^{\top} \cdot \diag (z(x)) \\
        - & ~ \alpha(y)^{-2} \cdot z(x) \cdot \wt{c}(x)^{\top} \cdot \diag (z(x)))\| \\
        \leq & ~ |\alpha(x)^{-2} - \alpha(y)^{-2} |\cdot \| z(x) \cdot \wt{c}(x)^{\top} \cdot \diag (z(x))\| \\
        \leq &~  4\beta^{-3} \sqrt{n}R \exp(R^2) \cdot \|x -y \|_2 \cdot \| z(x)\|_2 \cdot\|\wt{c}(x)^{\top}\|_2 \\
        \cdot & ~ \|z(x) \|_2 \\
        \leq & ~ 4\beta^{-3} \sqrt{n}R \exp(R^2) \cdot \|x -y \|_2\cdot  4 n \cdot \exp(2R^2) \cdot 10\sqrt{n} R_{f, 2}^2  \\
        \leq & ~ 160 \beta^{-3} n^2 \cdot  \exp(4R^2)\cdot (2 \sqrt{n} \beta^{-1} \exp(R^2))^2 \|x -y \|_2 \\
        \leq & ~ \beta^{-5} n^3 \cdot  \exp(7R^2) \|x -y \|_2
        \end{align*}
where the first step follows from definition of $G_{2,1}$, the second step follows from {\bf Part 6} of Fact~\ref{fac:matrix_norm}, the third step follows from {\bf Part 10} of Lemma~\ref{lem:basic_lips} and {\bf Part 7} of Fact~\ref{fac:matrix_norm}, the fourth step follows from {\bf Part 8, 10} of Lemma~\ref{lem:upper_bound}, the fifth step follow from the definition of $R_{f,2}$, and the last step follows from the simple algebra.

Let's prove $G_{2,2}$
\begin{align*}
    & ~ \|G_{2,2} \| \\
    = & ~ \| -( \alpha(y)^{-2} \cdot z(x) \cdot \wt{c}(x)^{\top} \cdot \diag (z(x)) \\
    - & ~ \alpha(y)^{-2} \cdot z(y) \cdot \wt{c}(x)^{\top} \cdot \diag (z(x)) )\| \\
    \leq & ~\|z(x) - z(y)\|_2 \cdot \|\alpha(y)^{-2}  \cdot \wt{c}(x)^{\top} \cdot \diag (z(x))  \|_2 \\
    \leq & ~ R \exp(R^2) \cdot  \|x- y \|_2 \cdot |\alpha(y)^{-2}| \cdot \| \wt{c}(x)^{\top}\|_2 \cdot \|z(x) \|_2 \\
    \leq & ~  R \exp(R^2) \cdot  \|x- y \|_2 \cdot \beta^{-2} \cdot 10\sqrt{n} R_{f, 2}^2 \cdot 2\sqrt{n}\exp(R^2)  \\
    \leq & ~  20 \beta^{-2} n \exp(3R^2) \cdot (2 \sqrt{n} \beta^{-1} \exp(R^2))^2 \|x -y \|_2 \\
    \leq & ~ \beta^{-4} n^2 \exp(6R^2) \|x -y \|_2
\end{align*}
where the first step follows from the definition of $G_{2,2}$, the second step follows from {\bf Part 9} of Fact~\ref{fac:vector_norm}, the third step follows from {\bf Part 7} of Lemma~\ref{lem:basic_lips} and {\bf Part 2, 4, and 7} of Fact~\ref{fac:vector_norm}, the fourth step follows from {\bf Part 8, 9, and 10} of Lemma~\ref{lem:upper_bound}, the fifth step follows from the definition of $R_{f,2}$ and Fact~\ref{fac:basic_algebraic_properties}, and the last step follows from the simple algebra.

Let's prove $G_{2,3}$
\begin{align*}
    \|G_{2,3} \| 
    = & ~ \| -(\alpha(y)^{-2} \cdot z(y) \cdot \wt{c}(x)^{\top} \cdot \diag (z(x)) \\
    - & ~ \alpha(y)^{-2} \cdot z(y) \cdot \wt{c}(y)^{\top} \cdot \diag (z(x)))\| \\
    \leq & ~ \|\wt{c}(x)^{\top} - \wt{c}(y)^{\top} \|_2 \cdot  \| \alpha(y)^{-2} \cdot z(y)  \cdot \diag (z(x))\|_2 \\
    \leq & ~ 5\sqrt{n}\cdot R_f \cdot R_{f,2}\cdot \|x -y \|_2 \cdot |\alpha(y)^{-2} | \cdot \| z(y) \|_2 \cdot \| z(x)\|_2 \\
    \leq & ~ 5\sqrt{n}\cdot R_f \cdot R_{f,2} \cdot  \beta^{-2} 4n\exp(2R^2) \cdot \|x -y \|_2  \\
    \leq & ~ \beta^{-5}n^3 \exp(7R^2) \| x- y\|_2 
\end{align*}
where the first step follows from the definition of $G_{2,3}$, the second step follows from {\bf Part 9} of Fact~\ref{fac:vector_norm}, the third step follows from {\bf Part 11} of Lemma~\ref{lem:basic_lips}, the fourth step follows from {\bf Part 8 and 9} of Lemma~\ref{lem:upper_bound}, and the last step follows from $R_f = 6  \beta^{-2} \cdot n \cdot  \exp(3R^2)$ and $R_{f, 2} = 2 \sqrt{n} \beta^{-1} \exp(R^2)$.

Let's prove $G_{2,4}$
\begin{align*}
    \|G_{2,4} \| 
    = & ~ \| -(\alpha(y)^{-2} \cdot z(y) \cdot \wt{c}(y)^{\top} \cdot \diag (z(x)) \\
    - & ~ \alpha(y)^{-2} \cdot z(y) \cdot \wt{c}(y)^{\top} \cdot \diag (z(y)))\| \\
    \leq & ~ \|\alpha(y)^{-2} \cdot z(y) \cdot \wt{c}(y)^{\top} \|\|\diag(z(x)) - \diag(z(y)) \| \\
    \leq & ~ |\alpha(y)^{-2} | \cdot \| z(y) \|_2 \cdot \| \wt{c}(x)^{\top}\|_2 \cdot R \exp(R^2) \cdot  \|x- y \|_2 \\
    \leq & ~  \beta^{-2} \cdot 2\sqrt{n} \cdot   \exp(R^2)\cdot 10 \sqrt{n} R_{f, 2}^2 \cdot  \|x- y \|_2  \\
    \leq & ~ \beta^{-4} n^2 \exp(4R^2)  \|x- y \|_2 
\end{align*}
where the first step follows from the definition of $G_{2,4}$, the second step follows from {\bf Part 4} of Fact~\ref{fac:matrix_norm}, the third step follows from {\bf Part 7} of Fact~\ref{fac:vector_norm} and {\bf Part 9} of Lemma~\ref{lem:basic_lips}, the fourth step follows from {\bf Part 8, 9, and 10} of Lemma~\ref{lem:upper_bound}, and the last step follows from $R_{f, 2} = 2 \sqrt{n} \beta^{-1} \exp(R^2)$.

Finally, by combining above results we can get
\begin{align*}
    \|G_{2}(x) - G_{2}(y) \|
    = & ~ \|G_{2,1} + G_{2,2} + G_{2,3} + G_{2,4} \| \\
    \leq & ~ \beta^{-5} n^3 \cdot  \exp(7R^2) \|x -y \|_2 \\
    + & ~ \beta^{-4} n^2 \exp(6R^2) \|x -y \|_2 \\
    + & ~  \beta^{-5}n^3 \exp(7R^2) \| x- y\|_2 \\
    + & ~  \beta^{-4} n^2 \exp(4R^2)  \|x- y \|_2  \\
    \leq & ~ 4 \beta^{-5}n^3\exp(7R^2) \cdot \|x -y \|_2
\end{align*}
    where the first step follows from the definitions of $G_{2,1}, G_{2,2}, G_{2,3}, G_{2,4}$, the second step follows from previous results, and the last step follows from simple algebra.
\end{proof}

\subsection{  Calculation: Step 3 Lipschitz for Matrix Function \texorpdfstring{$\alpha(x)^{-2} \cdot \diag (z(x)) \cdot \wt{c}(x) \cdot z(x)^{\top}$}{}}
\label{sub:hessian_lip:step3}

In this section, we analyze the third step, namely the Lipschitz for the matrix function $\alpha(x)^{-2} \cdot \diag (z(x)) \cdot \wt{c}(x) \cdot z(x)^{\top}$.

\begin{lemma}\label{lem:lips_G_3}
Let $G_3 = \alpha(x)^{-2} \cdot \diag (z(x)) \cdot \wt{c}(x) \cdot z(x)^{\top}$. 

    Then we have
    \begin{align*}
        \|G_3(x) - G_3(y)\| \leq 4 \beta^{-5}n^3\exp(7R^2) \cdot \|x -y \|_2
    \end{align*}
\end{lemma}
\begin{proof}

    The proof of $\|G_3(x) - G_3(y)\|$ is similar to $\|G_2(x) - G_2(y)\|$, so we omit it here.
\end{proof}

\subsection{  Calculation: Step 4 Lipschitz for Matrix Function \texorpdfstring{$\alpha(x)^{-1} \cdot \mathrm{diag} ( \wt{c}(x) \circ u_2(x) )$}{}}
\label{sub:hessian_lip:step4}

In this section, we analyze the fourth step, namely the Lipschitz for the matrix function $\alpha(x)^{-1} \cdot \mathrm{diag} ( \wt{c}(x) \circ u_2(x) )$.

\begin{lemma}\label{lem:lips_G_4}
Let $G_4 = \alpha(x)^{-1} \cdot \mathrm{diag} ( \wt{c}(x) \circ u_2(x) )$.

    Then we have
    \begin{align*}
        \| G_4(x) - G_4(y) \| \leq  2 \beta^{-4}  n^3 \cdot   \exp(7R^2) \|x -y \|_2 
    \end{align*}
\end{lemma}
\begin{proof}

We define
    \begin{align*}
        G_{4,1} 
        := & ~ \alpha(x)^{-1} \cdot \mathrm{diag} ( \wt{c}(x) \circ u_2(x) ) \\
        - & ~ \alpha(y)^{-1} \cdot \mathrm{diag} ( \wt{c}(x) \circ u_2(x) ) \\
        G_{4,2} 
        := & ~ \alpha(y)^{-1} \cdot \mathrm{diag} ( \wt{c}(x) \circ u_2(x) ) \\
        - & ~ \alpha(y)^{-1} \cdot \mathrm{diag} ( \wt{c}(y) \circ u_2(y) ) \\
    \end{align*}

Let's prove $G_{4,1}$ first,
\begin{align*}
     & ~ \|  G_{4,1} \|\\ 
     = & ~ \| \alpha(x)^{-1} \cdot \mathrm{diag} ( \wt{c}(x) \circ u_2(x) ) - \alpha(y)^{-1} \cdot \mathrm{diag} ( \wt{c}(x) \circ u_2(x) )\| \\
     \leq & ~ |\alpha(x)^{-1} - \alpha(y)^{-1}  | \cdot \|\diag ( \wt{c}(x) \circ u_2(x))\|\\
     \leq & ~ |\alpha(x)^{-1} - \alpha(y)^{-1}  | \cdot \|\diag ( \wt{c}(x)) \| \cdot \| \diag( u_2(x))\|\\
     \leq & ~ 4\beta^{-2}
      \cdot R \cdot n\exp(2R^2) \cdot \|x -y \|_2   \cdot \|\wt{c}(x) \|_2  \cdot \|  u_2(x)\|_2\\
    \leq & ~ 4\beta^{-2}
      \cdot R \cdot n\exp(2R^2) \cdot \|x -y \|_2   \cdot 10 \sqrt{n} R_{f, 2}^2 \sqrt{n}\exp(R^2) \\
      \leq & ~ \beta^{-4} n^3 \exp(7R^2) \|x -y \|_2
\end{align*}
where the first step follows from definition of $G_{4,1}$, the second step follows from {\bf Part 6} of Fact~\ref{fac:matrix_norm}, the third step follows from Fact~\ref{fac:circ_rules}, the fourth step follows from {\bf Part 4} of Lemma~\ref{lem:basic_lips} and {\bf Part 2 and 4} of Fact~\ref{fac:vector_norm}, the fifth step follows from {\bf Part 1, 10} of Lemma~\ref{lem:upper_bound}, and the last step follows from Fact~\ref{fac:basic_algebraic_properties} and $R_{f, 2} = 2 \sqrt{n} \beta^{-1} \exp(R^2)$.

Then let's prove $G_{4,2}$
\begin{align*}
    & ~ \|G_{4,2}\| \\
    = & ~ \| \alpha(y)^{-1} \cdot \mathrm{diag} ( \wt{c}(x) \circ u_2(x) ) - \alpha(y)^{-1} \cdot \mathrm{diag} ( \wt{c}(y) \circ u_2(y) )\| \\
    \leq & ~ \| \mathrm{diag} ( \wt{c}(x) \circ u_2(x) ) - \mathrm{diag} ( \wt{c}(y) \circ u_2(y) )\| |\alpha(y)^{-1} |\\
    \leq & ~  10 \beta^{-1} n \cdot R_f \cdot R_{f,2} \cdot \exp(2R^2)  \|x -y \|_2\\
    \leq & ~   \beta^{-4} n^{2.5} \cdot \exp(7R^2)  \|x -y \|_2
    \end{align*}
where the first step follows from the definition of $G_{4,2}$, the second step follows from {\bf Part 7} of Fact~\ref{fac:vector_norm}, and the second step follows from {\bf Part 12} of Lemma~\ref{lem:basic_lips} and {\bf Part 4} of Lemma~\ref{lem:upper_bound}, and the last step follows from $R_f = 6  \beta^{-2} \cdot n \cdot  \exp(3R^2)$ and $R_{f, 2} = 2 \sqrt{n} \beta^{-1} \exp(R^2)$.

By combining the above results, we can get
\begin{align*}
    & ~ \|G_4(x) - G_4(y) \|\\
    = & ~ \|G_{4,1} + G_{4,2} \|\\
    \leq & ~ \beta^{-4} n^3 \exp(7R^2) \|x -y \|_2 +  \beta^{-4} n^{2.5} \cdot \exp(7R^2)  \|x -y \|_2 \\
    \leq & ~ 2 \beta^{-4}  n^3 \cdot   \exp(7R^2) \|x -y \|_2 
\end{align*}
where the first step follows from the definitions of $G_{4,1}, G_{4,2}$, the second step follows from previous results, and the last step follows from simple algebra.
\end{proof}

\section{Main Result}
\label{sec:main_result}

\ifdefined\isarxiv

\else

In Section~\ref{sub:main_result:newton}, we introduce the background of the approximate version of Newton's method. In Section~\ref{sub:main_result:main}, we present our main algorithm and theorem.

\subsection{Background of Newton's Method}
\label{sub:main_result:newton}

In this section, we introduce the background of Newton's method.

\begin{definition}[$(l,M)$-good Loss function]\label{def:f_ass}
For a function $L : \R^d \rightarrow \R$, we say $L$ is $(l,M)$-good it satisfies the following conditions,
\begin{itemize}
    \item {\bf $l$-local Minimum.}  
    We define $l >0$ to be a positive scalar. If there exists a vector $x^* \in \R^d$ such that the following holds
    \begin{itemize}
        \item $\nabla L(x^*) = {\bf 0}_d$.
        \item $\nabla^2 L(x^*) \succeq l \cdot I_d$.
    \end{itemize}
    \item {\bf Hessian is $M$-Lipschitz.} If there exists a positive scalar $M>0$ such that
    \begin{align*}
        \| \nabla^2 L(y) - \nabla^2 L(x) \| \leq M \cdot \| y - x \|_2 
    \end{align*}
    \item {\bf Good Initialization Point.} Let $x_0$ denote the initialization point. If $r_0:=\| x_0 -x_*\|_2$ satisfies
    \begin{align*}
        r_0 M \leq 0.1 l
    \end{align*}    
\end{itemize}
\end{definition}

\begin{lemma}[\cite{dsw22,syyz22}]\label{lem:subsample}
Let $\epsilon_0 = 0.01$ be a constant precision parameter. 
Let $A \in \R^{n \times d}$ be a real matrix, then for any positive diagonal (PD) matrix $D \in \R^{n \times n}$, there exists an algorithm which runs in time
\begin{align*}
O( (\nnz(A) + d^{\omega} ) \poly(\log(n/\delta)) )
\end{align*}
and it outputs an $O(d \log(n/\delta))$ sparse diagonal matrix $\wt{D} \in \R^{n \times n}$ for which 
\begin{align*}
(1- \epsilon_0) A^\top D A \preceq A^\top \wt{D} A \preceq (1+\epsilon_0) A^\top D A.
\end{align*}
Note that, $\omega$ denotes the exponent of matrix multiplication, currently $\omega \approx 2.373$ \cite{w12,lg14,aw21}.
\end{lemma}

\subsection{Main result}
\label{sub:main_result:main}

\fi

Now, we present our main theorem and algorithm.

\begin{algorithm}[!ht]\caption{Main algorithm of solving the Soft-Residual Regression problem in Definition~\ref{def:our_regression}.}\label{alg:main}
\begin{algorithmic}[1]
\Procedure{IterativeSoftResidualRegression}{$A \in \R^{n \times d},b \in \R^n,w \in \R^n, \epsilon, \delta$} \Comment{Theorem~\ref{thm:main_formal}} 
    \State Choose $x_0$ (suppose it satisfies Definition~\ref{def:f_ass})
    \State We use $T \gets \log( \| x_0 - x^* \|_2 / \epsilon )$ to denote the number of iterations.
    \For{$t=0 \to T$} 
        \State $D \gets B_{\diag}(x_t) + \diag(w \circ w)$ 
        \State $\wt{D} \gets \textsc{SubSample}(D,A,\epsilon_1 = \Theta(1), \delta_1 = \delta/T)$ \Comment{Lemma~\ref{lem:subsample}}
        \State $g \gets A^\top (f(x_t) \langle c(x_t) , f(x_t) \rangle + \diag(f(x_t)) c(x_t) )$
        \State $\wt{H} \gets A^\top \wt{D} A$
        \State $x_{t+1} \gets x_t + \wt{H}^{-1} g$ 
    \EndFor
    \State $\wt{x}\gets x_{T+1}$
    \State \Return $\wt{x}$
\EndProcedure
\end{algorithmic}
\end{algorithm}

\begin{theorem}[Main theorem]\label{thm:main_formal}
Let $A$ be an arbitrary matrix in $\R^{n \times d}$. Let $b$ and $w$ be arbitrary vectors in $\R^n$. Let $f(x) = \langle \exp(Ax) + Ax, \mathbf{1}_n \rangle^{-1} (\exp(Ax) + Ax) \in \R^n$ be defined as in Definition~\ref{def:basic_functions}. Let $x^*$ as the optimal solution of 
\begin{align*}
\| \langle \exp(Ax) + A x , {\bf 1}_n \rangle^{-1} ( \exp(Ax) + Ax ) - b \|_2^2,
\end{align*}
for $g(x^*) = \nabla f(x^*) = {\bf 0}_d$ and $\| x^* \|_2 \leq R$, with $R > 4$. Suppose $\|A\| \leq R$, each entry of $b$ is greater than or equal to $0$, $\|b\|_1 \leq 1$, $w_{i}^2 \geq 100 + l/\sigma_{\min}(A)^2$ for all $i \in [n]$, and $M = n^{1.5} \exp(30R^2)$.

Let $x_0$ denote an initial point for which it holds that $M \| x_0 - x^* \|_2 \leq 0.1 l$.

Then for all accuracy parameter $\epsilon \in (0,0.1)$ and failure probability $\delta \in (0,0.1)$, there exists a randomized algorithm (Algorithm~\ref{alg:main}) such that, with probability at least $1-\delta$, it runs $T = \log(\| x_0 - x^* \|_2/ \epsilon)$ iterations and outputs a vector $\wt{x} \in \R^d$ such that
\begin{align*}
\| \wt{x} - x^* \|_2 \leq \epsilon,
\end{align*}
and the time cost per iteration is
\begin{align*}
O( (\nnz(A) + d^{\omega} ) \cdot \poly(\log(n/\delta)). 
\end{align*}

Here $\omega$ denotes the exponent of matrix multiplication. Currently $\omega \approx 2.373$ \cite{w12,lg14,aw21}. 
\end{theorem}
\begin{proof}
\ifdefined\isarxiv
The proof follows from Lemma~\ref{lem:subsample}, Lemma~\ref{lem:one_step_shrinking} and Lemma~\ref{lem:newton_induction}.
\else
The proof details can be found in full version \cite{full}.
\fi
\end{proof}

\section{Conclusion}
\label{sec:conclusion}

In this paper, we propose a unified scheme of combining the softmax regression and ResNet by analyzing the regression problem 
\begin{align*}
    \| \langle \exp(Ax) + A x , {\bf 1}_n \rangle^{-1} ( \exp(Ax) + Ax ) - b \|_2,
\end{align*}
where $A \in \R^{n \times d}$ and $b \in \R^n$. The softmax regression focuses on analyzing $\exp(Ax)$, and the ResNet focuses on analyzing $F(x) + x$. We combine these together and study $\exp(Ax) + Ax$. 

Specifically, we formally define this regression problem. We show that the Hessian matrix is positive semidefinite with the loss function $L(x)$. We analyze the Lipschitz properties and approximate Newton's method. Our unified scheme builds a connection between two previously thought unrelated areas in machine learning, providing new insight into the loss landscape and optimization for the emerging over-parametrized neural networks.

In the future, researchers may implement an experiment with the proposed unified scheme on large datasets to test our theoretical analysis. Moreover, extending the current analysis to multi-layer networks is another promising direction. We believe that our unified perspective between softmax regression and ResNet will inspire more discoveries at the intersection of theory and practice of deep learning.


\ifdefined\isarxiv
\bibliographystyle{alpha}
\bibliography{ref}
\else
\bibliography{ref}
\bibliographystyle{IEEEtran}

\fi

\ifdefined\isarxiv

\newpage
\onecolumn
\appendix

\section{Approximate Newton Method}\label{sec:newton}

In Section~\ref{sec:newton:definitions}, we introduce the basic definitions and the update rule. In Section~\ref{sec:newton:approximation}, we present the approximation of Hessian and update rule.

\subsection{Definition and Update Rule}\label{sec:newton:definitions}

In this section, we introduce the basic definitions. We focus on analyzing the following function:
\begin{align*}
    \min_{x \in \R^d } L(x).
\end{align*}

Here, we give the definition of $(l,M)$-good Loss function.

\begin{definition}[$l$-local Minimum]\label{def:local_min}

    Let $L : \R^d \rightarrow \R$ be a function.

    Let $l > 0$ be a positive real number. 

    There exists $x^* \in \R^d$ such that
    \begin{align*}
        \nabla L(x^*) = {\bf 0}_d
    \end{align*}
    and
    \begin{align*}
        \nabla^2 L(x^*) \succeq l \cdot I_d.
    \end{align*}
\end{definition}

\begin{definition}[Hessian is $M$-Lipschitz]\label{def:hessian_lipschitz}
    Let $L : \R^d \rightarrow \R$ be a function.

    Let $M > 0$ be a positive real number. 

    The Hessian of $L$ is $M$-Lipschitz if
    \begin{align*}
        \| \nabla^2 L(y) - \nabla^2 L(x) \| \leq M \cdot \| y - x \|_2 
    \end{align*}
\end{definition}

\begin{definition}[Good Initialization Point]\label{def:good_point}
    Let $L : \R^d \rightarrow \R$ be a function.

    Let $x_0 \in \R^d$ be the initialization point.

    Let $x^* \in \R^d$.

    Let $r_0 = \|x_0 - x^*\|_2 \in \R$.

    Let $M > 0$ be a positive real number. 

    If $r_0$ satisfy $r_0 M \leq 0.1 l$, then we say $x_0$ is a good initialization point .
\end{definition}

\begin{definition}[$(l,M)$-good Loss function]\label{def:f_ass}

Let $L : \R^d \rightarrow \R$ be a function.

$L$ is called a $(l,M)$-good function if it satisfies $l$-local Minimum (Definition~\ref{def:local_min}), Hessian is $M$-Lipschitz (Definition~\ref{def:hessian_lipschitz}), and good initialization point (Definition~\ref{def:good_point}).

\end{definition}

Then, we define the gradient and Hessian.

\begin{definition}[Gradient and Hessian]\label{def:gradient_hessian}
Let $L : \R^d \rightarrow \R$ be a function.

The function $g : \R^d \rightarrow \R^d$, which is defined as
\begin{align*}
    g(x) := \nabla L(x),
\end{align*}
is called the gradient of the function $L$.

The function $H : \R^d \rightarrow \R^{d \times d}$, which is defined as
\begin{align*}
    H(x) := \nabla^2 L(x),
\end{align*}
is called the Hessian of the function $L$.

\end{definition}

After that, we introduce the definition of the exact update of Newton's method.
\begin{definition}[Exact update of the Newton method]\label{def:exact_update_variant}
\begin{align*}
    x_{t+1} = x_t - H(x_t)^{-1} \cdot g(x_t)
\end{align*}
\end{definition}

\subsection{Approximate of Hessian and Update Rule}\label{sec:newton:approximation}

In this section, we introduce the approximate of Hessian and update rule. We define the approximate Hessian as follows.

\begin{definition}[Approximate Hessian]\label{def:wt_H}

Let $H$ be defined as in Definition~\ref{def:gradient_hessian}.

We define $\wt{H}: \R^d \rightarrow \R^{d \times d}$ to be a function satisfying:
\begin{align*}
 (1-\epsilon_0) \cdot H(x_t) \preceq \wt{H}(x_t) \preceq (1+\epsilon_0) \cdot H(x_t) .
\end{align*}
\end{definition}

$\wt{H}$ is the approximated Hessian. We apply Lemma~4.5 in \cite{dsw22} to get the approximated Hessian.

\begin{lemma}[\cite{syyz22,dsw22}]\label{lem:subsample}

Let $\epsilon_0 = 0.01 \in \R$, which is called the constant precision parameter. 

Let $A \in \R^{n \times d}$.

Let $D \in \R^{n \times n}$ be an arbitrary positive diagonal matrix.

There is an algorithm that can run in a time complexity of approximately
\begin{align*}
O( (\nnz(A) + d^{\omega} ) \poly(\log(n/\delta)) ).
\end{align*}

This algorithm produces a sparse diagonal matrix $\wt{D}$ which is in $\R^{n \times n}$.

The key property of is that it satisfies the following equation:
\begin{align*}
(1- \epsilon_0) A^\top D A \preceq A^\top \wt{D} A \preceq (1+\epsilon_0) A^\top D A.
\end{align*}

It is worth noting that $\omega$ represents the exponent related to matrix multiplication, with an approximate value of $\omega \approx 2.373$ (see \cite{lg14,w12,aw21}).

\end{lemma}

Following the standard in the literature on Approximate Newton Hessian, as in \cite{a00, jkl+20, bpsw21, szz21, hjs+22, lsz23}, we take into account the following.

\begin{definition}[Approximate update]\label{def:update_x_k+1}

We examine the following procedure:
\begin{align*}
    x_{t+1} = x_t  - \wt{H}(x_t)^{-1} \cdot  g(x_t)  .
\end{align*}
\end{definition}

We present a technique from previous research.
\begin{lemma}[Iterative shrinking Lemma, Lemma 6.9 of \cite{lsz23}]\label{lem:one_step_shrinking}

Let $L : \R^d \rightarrow \R$ be a $(l,M)$-good function, as defined in Definition~\ref{def:f_ass}.

Let $\epsilon_0$ be a positive real number in $(0,0.1)$.

Let $x_t, x^* \in \R^d$ and $r_t:= \| x_t - x^* \|_2 \in \R$.

Let $M > 0$ be a positive real number and $\ov{r}_t: = M \cdot r_t$.

Then, 
\begin{align*}
r_{t+1} \leq 2 \cdot (\epsilon_0 + \ov{r}_t/( l - \ov{r}_t ) ) \cdot r_t.
\end{align*}

\end{lemma}

To represent the total number of iterations of the algorithm, let's use the symbol $T$. In order to utilize Lemma~\ref{lem:one_step_shrinking}, we need the following induction hypothesis. This lemma is a well-established concept in the literature in \cite{lsz23}.

\begin{lemma}[Induction hypothesis, Lemma 6.10 on page 34 of \cite{lsz23}]\label{lem:newton_induction}

Let $i$ be an arbitrary element in $[t]$.

Let $x_i, x^* \in \R^d$ and $r_i:= \| x_i - x^* \|_2 \in \R$.

Let $\epsilon_0$ be a positive real number in $(0,0.1)$.

Let $0.4 \cdot r_{i-1} \geq r_{i}$.

Let $0.1 l \geq M \cdot r_i$, where $M > 0$.

Then,
\begin{itemize}
    \item $0.4 r_t \geq r_{t+1}$
    \item $0.1 l \geq M \cdot r_{t+1}$
\end{itemize}

\end{lemma}

\else

\fi




\end{document}